\documentclass[runningheads]{llncs}

\usepackage{eccv}

\newcommand{\singleappendix}[1]{%
  \appendix
  \section*{#1}
  \addcontentsline{toc}{section}{#1}
  \stepcounter{section}
}

\usepackage{diagbox}
\usepackage[dvipsnames]{xcolor}
\usepackage{siunitx}
\usepackage{eccvabbrv}
\usepackage{tikz}
\usepackage{pgfplots}
\DeclareUnicodeCharacter{2212}{−}
\usepgfplotslibrary{groupplots,dateplot}
\usetikzlibrary{patterns,shapes.arrows}
\usepackage{amsmath}

\DeclareMathOperator*{\argmax}{argmax}
\usepackage{wrapfig}
\usepackage{graphicx}
\usepackage{lipsum}  % for generating dummy text
\usepackage{wrapfig}
\usepackage[ruled,vlined]{algorithm2e}
\usepackage{colortbl}
\usepackage{graphicx}
\usepackage{subcaption}
\usepackage{xcolor}
\usepackage{float}
\usepackage{listings}
\definecolor{codegreen}{rgb}{0,0.6,0}
\definecolor{codegray}{rgb}{0.5,0.5,0.5}
\definecolor{codepurple}{rgb}{0.58,0,0.82}
\definecolor{backcolour}{rgb}{0.95,0.95,0.92}

\usepackage{amssymb}% http://ctan.org/pkg/amssymb
\usepackage{pifont}% http://ctan.org/pkg/pifont

\usetikzlibrary{patterns}
\usepackage{graphicx}
\usepackage{booktabs}
\usepackage{pgfplots}
\usepackage{multirow}
\usepackage{adjustbox}
\usepackage[accsupp]{axessibility}  % Improves PDF readability for those with disabilities.

\usepackage{bbm}
\usepackage{hyperref}

\usepackage{orcidlink}

\begin{document}

% ---------------------------------------------------------------
% TODO REVIEW: Replace with your title
% \title{Enhancing Exemplar-Free Semi-supervised Class Incremental Learning Using Foundation Models} 
\title{TACLE: Task and Class-aware Exemplar-free Semi-supervised Class Incremental Learning} 
%\title{TaLE: Task and class aware weighted loss for exemplar-free semi-supervised class incremental learning} 
% TODO REVIEW: If the paper title is too long for the running head, you can set
% an abbreviated paper title here. If not, comment out.
\titlerunning{TACLE for EFSS-CIL}

% TODO FINAL: Replace with your author list. 
% Include the authors' OCRID for the camera-ready version, if at all possible.
\author{Jayateja Kalla$^\ast$ \and
Rohit Kumar$^\ast$ \and
Soma Biswas}

% TODO FINAL: Replace with an abbreviated list of authors.
\authorrunning{J. Kalla et al.}
% First names are abbreviated in the running head.
% If there are more than two authors, 'et al.' is used.
%
\institute{Department of Electrical Engineering, \\Indian Institute of Science, Bangalore, India. \\
\email{\{jayatejak, krohit, somabiswas\}@iisc.ac.in}}

\maketitle
\def\thefootnote{$\ast$}\footnotetext{\scriptsize{contributed equally to this work}}

\begin{abstract}
We propose a novel TACLE (\textbf{TA}sk and \textbf{CL}ass-awar\textbf{E}) framework to address the relatively unexplored and challenging problem of exemplar-free semi-supervised class incremental learning. In this scenario, at each new task, the model has to learn new classes from both (few) labeled and unlabeled data without access to exemplars from previous classes. 
In addition to leveraging the capabilities of pre-trained models, TACLE proposes a novel task-adaptive threshold, thereby maximizing the utilization of the available unlabeled data as incremental learning progresses.
Additionally, to enhance the performance of the under-represented classes within each task, we propose a class-aware weighted cross-entropy loss. 
We also exploit the unlabeled data for classifier alignment, which further enhances the model performance. 
Extensive experiments on benchmark datasets, namely CIFAR10, CIFAR100, and ImageNet-Subset100 demonstrate the effectiveness of the proposed TACLE framework.
We further showcase its effectiveness when the unlabeled data is imbalanced and also for the extreme case of one labeled example per class.  

  \keywords{semi-supervised class incremental learning \and task-adaptive threshold \and class-aware weighted cross-entropy loss}
\end{abstract}

% *************************************
\section{Introduction}
\label{sec:intro}
Recently, incremental or continual learning~\cite{li2017learning_algo_logit_2,rebuffi2017icarl_algo_logit_1} has emerged as an important research direction due to its wide applicability, especially in real-time scenarios where models need to adapt to new data continuously~\cite{golab2003issues_data_continous1, gomes2017survey_data_continoius2}. 
It addresses the practical limitations of collecting all data at once~\cite{krempl2014open_storage} and potential privacy concerns~\cite{chamikara2018efficient_privacy} where only the model is accessible but not the training data. But the major challenge faced by neural network models when dealing with continuous data stream is catastrophic forgetting~\cite{goodfellow2013empirical}, where previously learned knowledge is overwritten as the model adapts to new information.
Among various settings in continual learning~\cite{van2019three_scenerios}, Class Incremental Learning (CIL)~\cite{rebuffi2017icarl_algo_logit_1, castro2018end_cil1} has gained popularity due to its wide applicability, where the model is initially trained on a set of base classes and is subsequently updated when new sets of classes becomes available (referred to as a \textit{task}). 
%In the literature, the process of adding a set of new classes to a model is commonly referred to as a \textit{task}.
Most of the existing CIL works assume that significant amounts of labeled data is available at each task, which is quite restrictive.
It is only recently that researchers have started to address the more challenging and realistic Semi-Supervised Class-Incremental Learning (SS-CIL)~\cite{wang2021ordisco_sscil1, boschini2022continual_ccic_sscil2, kang2023soft_nncil_sscil3, kalla2023generalized_sscil4}, where the model has access to a few labeled samples per task, while most of remaining training data is unlabeled. 
\begin{figure}[t]
    \centering
    \includegraphics[scale=0.75]{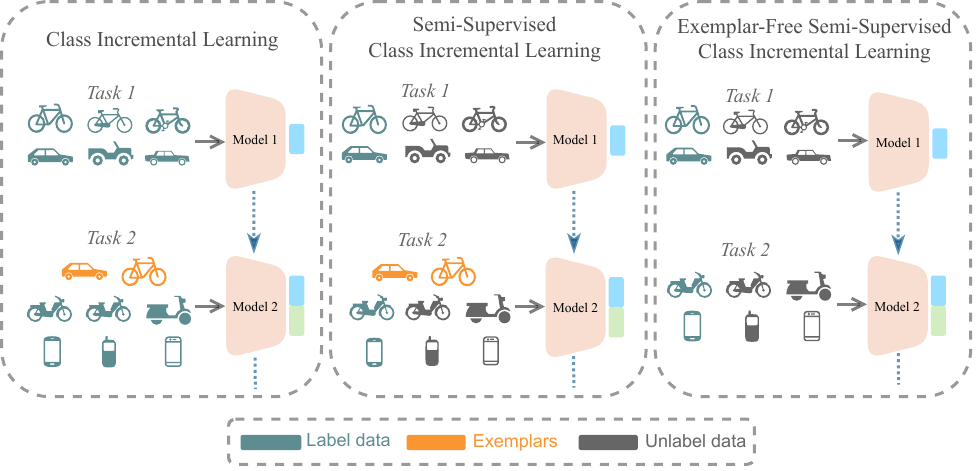}
    \caption{Difference between Class Incremental Learning (CIL), Semi-Supervised CIL (SS-CIL), and Exemplar-Free Semi-Supervised CIL (EFSS-CIL) settings.}
    \label{fig:cil_sscil_efsscil}
\end{figure}

In this work, we propose a novel framework termed TACLE (\textbf{TA}sk and \textbf{CL}ass awar\textbf{E}) for this challenging SS-CIL task, in a completely exemplar-free setting. Here, we do not need access to any examples of the previous classes, thereby complying to privacy concerns or requiring additional storage. Fig.~\ref{fig:cil_sscil_efsscil} shows the difference between CIL, SS-CIL and exemplar-free SS-CIL settings.
Inspired by the success of pre-trained models for many applications which also includes continual learning~\cite{hu2021well_pretraining, fini2022self_pretraining, mehta2023empirical_pretraining1, zhang2023slca}, we propose to leverage their generalization capacity for the challenging EFSS-CIL task. 
Specifically inspired by Slow Learner~\cite{zhang2023slca}, our approach is also a two-stage method. In the first stage, it learns robust feature representations, and in the second stage, it utilizes the mean and variance estimated from these features to adjust classifiers. For these two stages, we propose three novel modules: (i) a \textit{task-wise adaptive threshold} to effectively utilize unlabeled data across tasks; (ii) a \textit{class-aware weighted cross-entropy loss} to enhance the performance of under-represented classes by addressing the imbalance in the unlabeled data that surpasses the threshold, in the first stage; and (iii) exploiting the unlabeled data for better classifier alignment in the second stage, which further enhances the EFSS-CIL.
Extensive experiments conducted on various datasets, including CIFAR-10, CIFAR-100, and ImageNet-Subset100, demonstrate the effectiveness of the proposed modules in handling the EFSS-CIL task. 
To this end, our contributions are as follows: 
\begin{enumerate}
    \item To the best of our knowledge, TACLE is the first work to address the challenging EFSS-CIL setting. 
    \item We leverage pre-trained models to address EFSS-CIL, proposing three novel components: task adaptive threshold, class-aware weighted cross-entropy, and exploiting unlabeled data for classifier alignment.
    \item Extensive experiments on various datasets designed for EFSS-CIL demonstrate the effectiveness of the proposed approach. 
    \item Furthermore, experiments on extreme cases such as 1-shot EFSS-CIL and imbalanced scenarios further validate the efficacy of our framework.
\end{enumerate}

The rest of the paper is organized as follows: Section~\ref{sec:related_works} briefly overviews related works, Section~\ref{sec:problem_def} introduces the notations, while Section~\ref{sec:method} discusses the proposed methodology, and Section~\ref{sec:experiments} provides the experimental results. The paper concludes with an ablation study and analysis of the proposed components.
% **************************************
\section{Related Works}
\label{sec:related_works}
In this section, we briefly discuss about the related works in literature.\\
\textbf{1. Class Incremental Learning (CIL):} aims to progressively incorporate new classes into the model and multiple works have been proposed in literature to address the issues in CIL. These works can be broadly classified into three categories: \textbf{(i) Data-centric approaches:} These approaches~\cite{lopez2017gradient_gem_data_centric1, chaudhry2018efficient_agem_data_centric2, chaudhry2018riemannian_reimAN_walk_data_centric3, isele2018selective_experience_data_centric4, chaudhry2019tiny, buzzega2020dark,prabhu2020gdumb} mainly concentrate on adding exemplars to the model to alleviate catastrophic forgetting. However, their performance may degrade if no storage buffer is available for exemplars. 
Thus, many approaches~\cite{zhu2021prototype_pass_exemplar_free_cil1, zhu2022self_exemplar_free_cil2, petit2023fetril_exemplar_free_cil3, zhang2023slca} have started addressing the more realistic exemplar-free CIL setting, without access to exemplars from previous tasks.
Recently, there is a growing trend in leveraging pre-trained models for CIL~\cite{hu2021well_pretraining, fini2022self_pretraining, mehta2023empirical_pretraining1, zhang2023slca, wang2022dualprompt_model_centric_5, wang2022learning_model_centric_4}.
Our work is inspired from the recent SLCA~\cite{zhang2023slca} framework, which achieves impressive performance for exemplar-free CIL task. 
% *******************************
\textbf{(ii) Model-centric approaches:} These approaches~\cite{yoon2017lifelong_model_centric1, yan2021dynamically_model_centric2, rusu2016progressive_model_centric_3, wang2022learning_model_centric_4, wang2022dualprompt_model_centric_5, smith2023coda_model_centric_6} rely on dynamic expansion of the model to enhance their representation ability to mitigate catastrophic forgetting. 
Another line of approaches~\cite{chaudhry2018riemannian_reimAN_walk_data_centric3, zenke2017continual_SI_reg_1, aljundi2018memory_reg_2, kirkpatrick2017overcoming_reg_3} estimate the importance of weights in the model and apply regularization to those weights. 
\textbf{(iii) Algorithm-centric approaches:} focus on designing training strategies, such as knowledge distillation~\cite{hinton2015distilling_kd_hinton}, to mitigate catastrophic forgetting. 
%Knowledge distillation is used for transferring knowledge from teacher to student models and  
Knowledge distillation is used for transferring knowledge from old tasks to new tasks and different variants have been proposed: logits distillation~\cite{rebuffi2017icarl_algo_logit_1, li2017learning_algo_logit_2, hou2018lifelong_algo_logit_3, wu2019large}, feature distillation~\cite{hou2019learning_algo_feat_1, dhar2019learning_algo_feat_2, kang2022class_algo_feat_3, douillard2020podnet_algo_feat_4}, relation distillation~\cite{gao2022r_algo_relation_1, yu2020semantic_algo_relation_2}, etc. \\ \\
% *******************************
\textbf{2. Semi-Supervised Learning (SSL):} The SSL paradigm aims to train models effectively using a combination of labeled and unlabeled data. 
One of the successful and commonly used SSL frameworks involves consistency regularization and leveraging pseudo-labels for the unlabeled samples. 
FixMatch~\cite{sohn2020fixmatch} is a popular SSL technique that assigns pseudo-labels based on a predefined confidence threshold. 
%Our task-wise dynamic threshold draws inspiration from FixMatch but is tailored for incremental learning scenarios. 
Few other successful SSL approaches are ~\cite{berthelot2019mixmatch_ssl6, berthelot2019remixmatch_ssl7, li2020dividemix_ssl8}, which align the unlabeled data feature distributions and ~\cite{zhang2021flexmatch_ssl2, wang2022freematch_ssl3, chen2023softmatch_ssl4, chen2023boosting_ssl5}, which uses different pseudo-labeling strategies.
%Other variations using mix-up~\cite{zhang2017mixup} technique are proposed~\cite{berthelot2019mixmatch_ssl6, berthelot2019remixmatch_ssl7, li2020dividemix_ssl8} to align distributions and extensions of pseudo-labeling strategies have been proposed in ~\cite{zhang2021flexmatch_ssl2, wang2022freematch_ssl3, chen2023softmatch_ssl4, chen2023boosting_ssl5}.
\\ \\
% *******************************
\textbf{3. Semi-Supervised Class Incremental Learning (SS-CIL):} Online Replay with Discriminator Consistency (ORDisCo)~\cite{wang2021ordisco_sscil1} is a pioneering work addressing SS-CIL, focusing on interdependently learning a classifier with a conditional Generative Adversarial Network (GAN). This approach involves the continual transmission of the learned data distributions to the classifier. However, the method incurs prohibitive costs when applied to higher-resolution images like ImageNet-100. Boschinia et al.~\cite{boschini2022continual_ccic_sscil2} introduced Contrastive Continual Interpolation Consistency (CCIC) for this task, combining the advantages of rehearsal based methods with consistency regularization and distance-based constraints. In ESPN~\cite{kalla2023generalized_sscil4}, outliers are introduced in the unlabeled data to enhance the realism of the problem. More recently, NNCSL~\cite{kang2023soft_nncil_sscil3} proposed a soft nearest-neighbor framework to learn powerful and stable representations. 
In contrast, the proposed TACLE framework leverages pre-trained models to enhance representations, thereby eliminating the need for exemplars.

% **************************************
\section{Problem Formulation}
\label{sec:problem_def}
We now formally define the problem of Semi-Supervised Class Incremental Learning (SS-CIL) and introduce relevant notations used throughout the paper. 
In CIL, the model is trained on total $\mathcal{T}$ sequential data streams (or tasks) denoted by $\{\mathcal{D}^{(1)}, \mathcal{D}^{(2)},\ldots,\mathcal{D}^{(\mathcal{T})}\}$, each with its corresponding class label set is denoted by $\{\mathcal{C}^{(1)}, \mathcal{C}^{(2)},\ldots, \mathcal{C}^{(\mathcal{T})}\}$. 
Throughout this training process, the number of parameters in the feature extractor $\{\Theta\}$ remains unchanged. However, new classifiers $\{\psi^{(1)}, \psi^{(2)}, \ldots, \psi^{(\mathcal{T})}\}$ are incrementally added after training each task. In traditional CIL, at task $t$, the model have access to large amount of labeled data $\mathcal{D}^{(t)}$, where $t = 1,\ldots, \mathcal{T}$ along with old task exemplars.

In contrast, in SS-CIL, the data for the present task $t$ consists of both labeled and unlabeled samples i.e $\mathcal{D}^{(t)} \in \{ \mathcal{D}_{l}^{(t)} \cup \mathcal{D}_{ul}^{(t)} \}$. 
Here, it is assumed that both labeled and unlabeled samples come from the same task classes $\mathcal{C}^{(t)}$, and the number of labeled samples is significantly smaller compared to that of unlabeled data i.e.,  $|\mathcal{D}_{l}^{(t)}| << |\mathcal{D}_{ul}^{(t)}|$. Once the model $\{\Theta, \psi^{(1:t)}\}$ has learnt from the current task data $\mathcal{D}^{(t)}$, it has to perform well on all the classes seen so far i.e $\{C^{(1:t)}\}$. Throughout SS-CIL, the model learns one base task and a total of $\mathcal{T} -1$ tasks in an incremental fashion, with no overlap in the label space between the different tasks, i.e. $C^{(i)} \cap C^{(j)} = \phi$ for ($i \neq j$).  In the SS-CIL protocol, an exemplar bank $\mathcal{E}$ will be updated after each task to alleviate the catastrophic forgetting. However, in EFSS-CIL, there are no exemplars saved for future tasks (because of privacy or storage costs), which makes the problem more challenging and realistic.

% **************************************
\section{Proposed Method}
\label{sec:method}
Now, we describe in detail, our proposed TACLE (\textbf{TA}sk and \textbf{CL}ass-awar\textbf{E}) framework, designed specifically for EFSS-CIL. As discussed earlier, we have access to both labeled data $\mathcal{D}_{l}^{(t)} = \{\mathbf{x}_{i}^{l}, y_{i}^{l}\}_{i=1}^{N_{l}^{(t)}}$ and unlabeled data $\mathcal{D}_{ul}^{(t)} = \{\mathbf{x}_{i}^{ul}\}_{i=1}^{N_{ul}^{(t)}}$, for task $t$.  Here, $N_{l}^{(t)}, N_{ul}^{(t)}$ are the number of labeled and unlabeled samples, respectively. The TACLE framework adopts a two-stage training strategy for each task, namely (i) stage 1: \textit{Feature Representation Learning}: This stage leverages both labeled and unlabeled data to learn robust feature representations and (ii) stage 2: \textit{Classifier Alignment}: This stage focuses on aligning the classifiers with the learned features from both labeled and unlabeled data.
At task $t$, the model is trained by utilizing labeled data $\mathcal{D}_{l}^{(t)}$ through standard supervised loss: 
\begin{equation}
   \mathcal{L}_{s}(\mathbf{x}_{i}^{l}, y_{i}^{l}) =  \mathcal{H}(p_{i}^{l}, y_{i}^{l}) 
   \label{eq_ls}
\end{equation}
where $\mathbf{x}_{i}^{l}$ is the labeled sample and 
$p_{i}^{l} = \psi^{(t)}(\Theta(\mathbf{x}_{i}^{l}))$ is the predicted probability distribution given by the model for task $t$; $\mathcal{H}$ represents the standard cross-entropy loss.
Now, we describe the different proposed modules to effectively utilize the available unlabeled data at current task.

\subsection{Stage 1: Learning Feature Representations} 
\textbf{Task-wise adaptive threshold:} To leverage information from unlabeled data, we draw inspiration from SSL framework FixMatch~\cite{sohn2020fixmatch}, where, unlabeled data contributes to the learning process if the model's confidence surpasses the predefined threshold $\gamma$ (typically set to 0.95). 
In the EFSS-CIL setting, a fixed threshold across tasks may not be effective, since the number of classes increases with each task, thereby impacting the confidence scores of the unlabeled data. 
\begin{figure}[t]
% \vspace{-1.0cm}
\centering
\begin{tikzpicture}

    \definecolor{color0}{HTML}{DC5C52} %RED
    % \definecolor{color1}{HTML}{E6BF9E} %Orange
    \definecolor{color1}{HTML}{ddbea9} %Orange
    
    \definecolor{grid_color}{HTML}{e2e2e2} %grey
    \pgfplotsset{grid style={densely dashed, grid_color}}

    \begin{groupplot}[group style={group name = MoMP of Different datasets, group size=2 by 1,horizontal sep = 1.5cm}, height=4cm,width=6cm ]

        \nextgroupplot[xlabel={Tasks}, ylabel={ACS}, 
                   x label style={at={(axis description cs:0.5,0.05)},anchor=north},
                   y label style={at={(axis description cs:0.05,0.5)},anchor=north},
                   xmin=0, xmax=5.25, xtick={1,2,3,4,5}, 
                   xticklabels={1,2,3,4,5,6,7,8,9,10}, 
                   ymin=0.5, ymax=1.1, ytick={0.6,0.7,0.8,0.9,1}, 
                   title={a.) CIFAR10},
                   title style={anchor=north, yshift=2.5ex},
                   grid=major,]

            \addplot[densely dashed, color= color1, mark=*, mark options={solid}, mark size =2.3]  coordinates {
                        (1, 0.9863)
                        (2, 0.8642)
                        (3, 0.81)
                        (4, 0.83)
                        (5, 0.64)
                    };

        \nextgroupplot[xlabel={Tasks}, ylabel={ACS}, 
                   x label style={at={(axis description cs:0.5,0.05)},anchor=north},
                   y label style={at={(axis description cs:0.05,0.5)},anchor=north},
                   xmin=0, xmax=10.5, xtick={2,4,6,8,10}, 
                   xticklabels={2,4,6,8,10}, 
                   ymin=0.5, ymax=1.1, ytick={0.6,0.7,0.8,0.9,1}, 
                   title={b.) CIFAR100},
                   title style={anchor=north, yshift=2.5ex},
                   grid=major]
% {1,2,3,4,5,6,7,8,9,10}

                    \addplot[densely dashed, color= color1, mark=*, mark options={solid}, mark size =2.3]  coordinates {
                        (1, 0.9406)
                        (2, 0.80)
                        (3, 0.81)
                        (4, 0.78)
                        (5, 0.75)
                        (6, 0.67)
                        (7, 0.7)
                        (8, 0.69)
                        (9, 0.72)
                        (10, 0.62)
                    };

    \end{groupplot}
\end{tikzpicture} 
% \vspace{-0.3cm}

\caption{Illustrates the Average Confidence Score (ACS) for unlabeled data across tasks. The ACS calculated by taking average of maximum probability confidence scores from all the unlabeled data, at the end of training. The observed decaying trend indicates that using a fixed high threshold in SS-CIL may not be suitable for effective utilization of unlabeled data in feature learning. Due to the fixed threshold, the amount of unlabeled data utilized for training is significantly reduced as tasks progresses.}
\label{fig:momp_vs_tasks}
\end{figure}
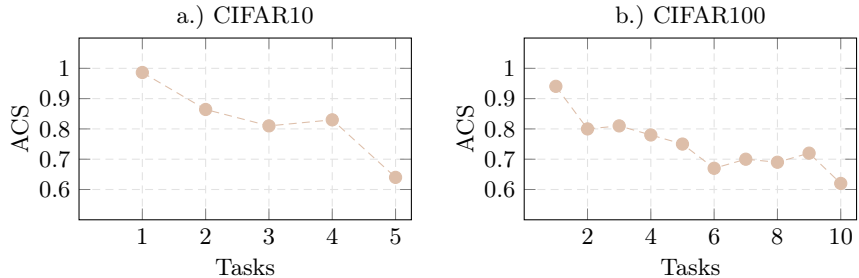

To analyze this confidence scores across tasks, we plot the Average Confidence Score (ACS) of the unlabeled data for CIFAR10 and CIFAR100 datasets after each task in Fig.~\ref{fig:momp_vs_tasks}.
The exact task splits are discussed in the experimental section. 
For ACS calculation, after training each task, we pass the respective task's unlabeled data through the model and calculate the average of their confidence scores, which provides insight into the average maximum confidence of the unlabeled data. 
Empirically, we observe that the ACS value reduces as the tasks progresses. As the number of classes increase with more tasks, it induces more confusion in the model predictions, thereby reducing the confidence value on the unlabeled data. 
To address this issue, we propose a \textit{task-wise adaptive threshold} instead of a fixed threshold to effectively leverage the unlabeled data available at task $t$. 
%Based on this observation we propose a \textit{task-wise adaptive threshold} to effectively utilize the unlabeled data. 

We denote a given unlabeled sample and its augmentation as $\mathbf{x}_{i}^{ul}, \mathbf{\hat{x}}_{i}^{ul}$, and their respective prediction probabilities as $p_{i}^{ul}, \hat{p}_{i}^{ul}$. 
The unsupervised loss, that incorporates a task-wise adaptive threshold is calculated as 
\begin{equation}
    \mathcal{L}_{us} (\mathbf{x}_{i}^{ul}) = \mathbbm{I}(\max(p_{i}^{ul})> \gamma_{a}^{(t)})  \cdot \mathcal{H}(\hat{p}_{i}^{ul},\argmax(p_{i}^{ul}))
    \label{eq_lus}
\end{equation}
Here, $\mathbb{I}$ is an indicator function which is 1 if the maximum value of model output probability $p_{i}^{ul}$ surpasses this adaptive threshold $\gamma_{a}^{(t)}$, otherwise, the loss is 0. The task-wise adaptive threshold, $\gamma_{a}^{(t)}$ is inspired from the inverse sigmoid function~\cite{menon1996characterization_inverse_sigmoid_1, dombi2022generalized_inverse_sigmoid_2}, and here we adapted it for the EFSS-CIL task as follows:
\begin{equation}
\gamma_{a}^{(t)} = \frac{\alpha}{1+e^{\alpha t}} + \beta,
\label{eq_task_adap_th}
\end{equation}
\begin{figure}[t]
    \centering
    \includegraphics[scale=0.8]{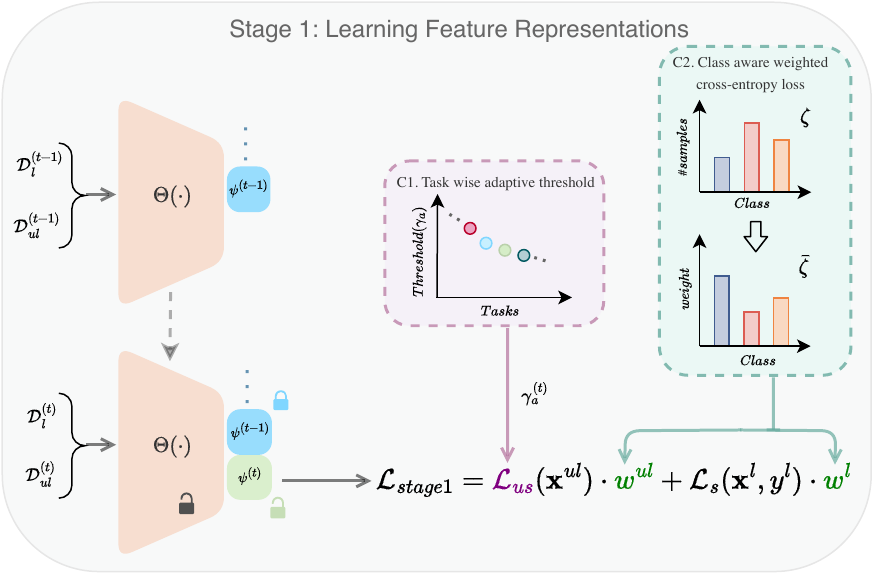}
    \caption{The proposed TACLE introduce two components in stage 1 training at task $t$: C1. Task-wise adaptive threshold ($\gamma_{a}^{(t)}$) is employed in the computation of the unsupervised loss $\mathcal{L}_{us}$. C2. Class-aware weights are utilized in the computation of both supervised and unsupervised losses, where the weights are determined based on the class-wise distribution of pseudo-unlabeled data.}
    \label{fig:stage1}
\end{figure}
We observe that the dynamic threshold computed using the above equation decreases as the task index $t$ increases, which aligns with the inverse sigmoid behavior. The hyper-parameters $\alpha$ and $\beta$ provide flexibility in controlling the rate of threshold reduction. This dynamic adjustment ensures an effective utilization of unlabeled data in the feature learning process, allowing the model to better adapt to different tasks. In all our experiments, across all datasets, we use $\alpha=0.5, \:\beta=0.65$. Further analyses of these choices and dynamic threshold behavior plots across tasks are provided in the Appendix. \\ \\
% ****************************
\textbf{Class-aware weighted loss:} While the task-wise adaptive threshold helps to learn better feature representations from unlabeled data across tasks, even within a task, there is significant class imbalance among samples surpassing the task-wise adaptive threshold. 
This imbalance can bias the model training towards classes with more pseudo-labels, hindering the performance on under-represented classes. To mitigate this, inspired from the SSL~\cite{zhang2021flexmatch_ssl2, chen2023boosting_ssl5} works, we propose a very simple, yet effective class-aware weighted cross-entropy loss.

At task $t$, after each epoch during stage 1, we calculate the normalized histogram of confident unlabeled samples across different classes. 
This histogram, represented as a vector $\zeta \in \mathbb{R}^{|\mathcal{C}^{(t)}|}$, serves as the basis for the class-aware weighted distribution $\bar{\zeta}$ used in the weighted cross-entropy loss calculation. The class-aware weighted distribution $\bar{\zeta}$ is calculated as $\bar{\zeta} = 2-\zeta$, which ensures that the class having maximum number of confident unlabeled samples in histogram $\zeta$ has $\bar{\zeta} = 1$, and class with least confident unlabeled samples has $\bar{\zeta} = 2$ ($1 \leq \bar{\zeta} \leq 2$). Essentially, this class-aware weighted distribution assigns higher weights to under-represented classes and lower weights to well-represented ones. 
Using this distribution $\bar{\zeta}$, we assign the weight $w_{i}^{l} = \bar{\zeta}_{y_{i}^{l}}$ for a labeled sample pair $(\mathbf{x}_{i}^{l}, y_{i}^{l})$. Similarly, for an unlabeled sample $\mathbf{x}_{i}^{ul}$, we set $w_{i}^{ul} = \bar{\zeta}_{\argmax(p_{i}^{ul})}$ ($w_{i}^{ul}$ is determined based on the pseudo-label i.e., $\argmax(p_{i}^{ul})$).
The total stage1 training loss incorporating these class-aware information calculated as
\begin{equation}
    \mathcal{L}_{stage1} = \mathcal{L}_{s}(\mathbf{x}_{i}^{l}, y_{i}^{l}) \cdot w_{i}^{l} + \mathcal{L}_{us}(\mathbf{x}_{i}^{ul}) \cdot w_{i}^{ul}
    \label{eq_stage1}
\end{equation}
Fig.~\ref{fig:stage1} illustrates the complete stage 1 training of TACLE, which utilizes the task-wise adaptive threshold and class-aware weighted loss to train the model for EFSS-CIL.
% \color{red} Based on this distribution $\bar{\zeta}$, for labeled sample pair $(\mathbf{x}_{i}^{l}, y_{i}^{l})$, we set $w_{i}^{l} = \bar{\zeta}_{y_{i}^{l}}$ and for an unlabeled sample $\mathbf{x}_{i}^{ul}$, we set $w_{i}^{ul} = \bar{\zeta}_{\argmax(p_{i}^{ul})}$, Here, the weight is based on the pseudo label i.e. $\argmax(p_{i}^{ul})$. \color{black}
% ***********************
\begin{figure}[t]
    \centering
    \includegraphics[scale=0.8]{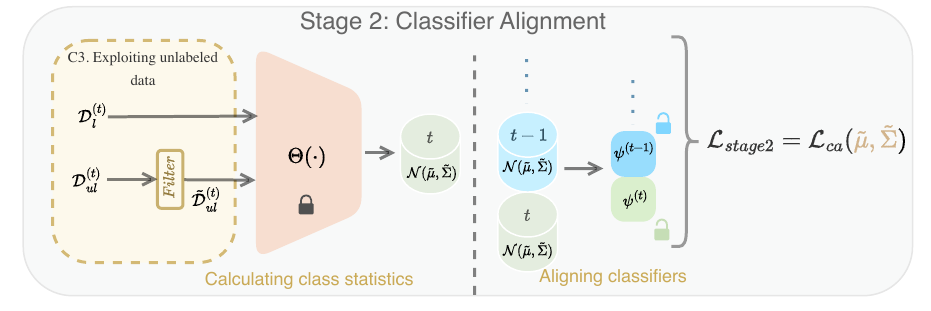}
    \caption{After stage 1 training, we filter out under-confident samples and create the expanded label set $\tilde{\mathcal{D}}^{(t)} = \mathcal{D}_{l}^{(t)} \cup \tilde{\mathcal{D}}_{ul}^{(t)}$. We estimate class statistics for task $t$ using this expanded label set. Utilizing class-wise statistics for all encountered classes, we fine-tune all classifiers with the classifier alignment loss $\mathcal{L}_{ca}$, defined in Eq.~\ref{eq_l_ca}. This comprehensive strategy which effectively utilizes the unlabeled data, constitutes our third component (C3) in the proposed approach.}
    \label{fig:stage2}
\end{figure}
\subsection{Stage 2: Classifier alignment using unlabeled data} 
In pre-trained models, aligning classifiers with the underlying class distributions plays a critical role in achieving optimal performance. In the SLCA method, classifier alignment involves utilizing class means and variances, denoted as $\{ \mu_{k}^{(t)}, \Sigma_{k}^{(t)}\}_{k=1}^{|\mathcal{C}^{(t)}|}$, calculated in feature space of dimension $d$, where $|\mathcal{C}^{(t)}|$ represents the number of classes in task $t$. These class distribution parameters $\mu_{k}^{(t)} \in \mathbb{R}^{d}$ and $\Sigma_{k}^{(t)} \in \mathbb{R}^{d \times d}$ are estimated from the available labeled data $\mathcal{D}_{l}^{(t)}$ and stored, along with the old task classes distributions (${ \mu_{k}^{(1:t-1)}, \Sigma_{k}^{(1:t-1)}}$). In stage 2 classifier alignment process, all the class distributions (${ \mu_{k}^{(1:t)}, \Sigma_{k}^{(1:t)}}$) from task 1 to $t$  are utilized to align all the classifiers in the model. For this purpose, a class-wise distribution is approximated by as a multi-dimensional Gaussian $\mathcal{N}(\mu_{k}^{(1:t)}, \Sigma_{k}^{(1:t)})$ function, from which features are sampled to align the classifiers of both the current task and all the previous tasks'. The classifier alignment loss is given by $\mathcal{L}_{ca}(\mu_{k}^{(1:t)}, \Sigma_{k}^{(1:t)}) = \mathcal{H}(\psi^{(1:t)}(z), k)$, where $z \sim \mathcal{N}(\mu_{k}^{(1:t)}, \Sigma_{k}^{(1:t)})$ are samples in feature space from all the classes seen so far. 

However, relying solely on labeled data might not accurately capture the true class distributions due to the inherent scarcity of labeled data in EFSS-CIL settings, specially if we have as few as a single labeled sample per class. 
Towards this goal, we propose to incorporate the confident unlabeled samples to better estimate of the class distributions parameters. We show that this can further aid the classifier alignment process.

 We achieve this by constructing an expanded label set, denoted by $\tilde{\mathcal{D}}^{(t)} = \mathcal{D}_{l}^{(t)} \cup \tilde{\mathcal{D}}_{ul}^{(t)}$.
 This combines the original labeled data with pseudo-labeled data derived from confident unlabeled samples defined as
%Here, $\tilde{\mathcal{D}}_{ul}^{(t)}$ comprises image and pseudo-label pairs from confident samples in the unlabeled data, defined as 
$\tilde{\mathcal{D}}_{ul}^{(t)} = \{\{\mathbf{x}_{i}^{ul}, \argmax{(p_{i}^{ul})}\} \mid \mathbf{x}_{i}^{ul} \in \mathcal{D}_{ul}^{(t)}, \max{(p_{i}^{ul})} > \gamma_{a}^{(t)}, i=1\ldots N_{ul}^{(t)}$\}. 
% Here, $\gamma$ is a predefined threshold on the model's predicted probability, 
%These confident samples are identified based on a predefined threshold $\gamma$ on the model's predicted probability, and the expanded label set is constructed. 
The improved statistics are calculated using $\tilde{\mathcal{D}}^{(t)}$ is denoted as $\{ \tilde{\mu}_{k}^{(t)}, \tilde{\Sigma}_{k}^{(t)}\}$ which are then utilized for classifier alignment in the stage 2 training loss function:
\begin{equation}
    \mathcal{L}_{stage2} = \mathcal{L}_{ca}(\tilde{\mu}_{k}^{(1:t)}, \tilde{\Sigma}_{k}^{(1:t)})
    \label{eq_l_ca}
\end{equation}
Fig.~\ref{fig:stage2} illustrates the stage 2 training process. These two stages are the same for each incremental task and these final aligned classifiers are used for classification during inference. Algorithm~\ref{alg:tacle} summarizes the TACLE training strategy for EFSS-CIL paradigm.

\begin{algorithm}[t]
% \SetKwComment{Comment}{/* }{ */}
\caption{TACLE for semi-supervised class incremental learning}
 \SetKwComment{Comment}{\qquad }{}
\KwIn{$\{\Theta, \psi\} \leftarrow $ Model; 
$\{\mathcal{D}^{(1)}, \mathcal{D}^{(2)},\ldots,\mathcal{D}^{(\mathcal{T})}\} \leftarrow $ Data stream;}
$E_{s1} \leftarrow$ No. of epochs for stage 1; $E_{s2} \leftarrow$ No. of epochs for stage 2;

\For{$t \leftarrow 1$ to $\mathcal{T}$}{
  % $\mathcal{D}^{t} \in {\mathcal{D}_{l}^{(t)} \cup  \mathcal{D}_{ul}^{(t)}}$\\
  $\mathcal{D}_{l}^{(t)} = \{\mathbf{x}_{i}^{l}, y_{i}^{l}\}_{i=1}^{N_{l}^{(t)}}; \mathcal{D}_{ul}^{(t)} = \{\mathbf{x}_{i}^{ul}\}_{i=1}^{N_{ul}^{(t)}};$

$\zeta \leftarrow \text{Uniform distribution across all classes}$

  \textcolor[HTML]{bc5a45}{\textit{$//$ $\# Stage $ 1: {Feature Representation Learning} $//$ }} \\
  \For{$e_{s1} \leftarrow 1$ to $E_{s1}$}{
    $\mathcal{B}_{l} = \text{SampleMiniBatch}(\mathcal{D}_{l}^{(t)})$;
    $\mathcal{B}_{ul} = \text{SampleMiniBatch}(\mathcal{D}_{ul}^{(t)})$;
    
    $\hat{\mathcal{B}}_{ul} = \text{ImageAugmentations}(\mathcal{B}_{ul})$;

     $\mathcal{O}_{l}, \mathcal{O}_{ul}, \hat{\mathcal{O}}_{ul} = \Theta(\psi^{(t)}(\mathcal{B}_{l}, \mathcal{B}_{ul}, \hat{\mathcal{B}}_{ul}));$

    $w^{l} \leftarrow \text{Assigning class-aware weights for labeled data }\mathcal{B}_{l} \text{ using } \bar{\zeta};$ 

    $w^{ul} \leftarrow \text{Assigning class-aware weights for unlabeled data }\mathcal{B}_{ul}\text{ using } \bar{\zeta};$ 
   
     $\mathcal{L}_{stage1} \leftarrow \mathcal{L}_{s}(\mathcal{B}_{l}) \cdot w^{l} + \mathcal{L}_{us}(\hat{\mathcal{B}}_{ul}) \cdot w^{ul};\:$ {\scriptsize{\tcp{Total loss for stage1 (Eq. ~\ref{eq_stage1})}}} 
     % \hspace{5.2cm}{\scriptsize{\tcp{$\mathcal{L}_{us} \text{ uses task wise threshold } \gamma_{a}^{t}$ \text{ (Eq.~\ref{eq_lus})}}}}\\

     $\zeta \leftarrow \text{Update the histogram distribution using }\mathcal{D}_{ul}^{(t)}, \gamma_{a}^{(t)};$

     $\bar{\zeta} \leftarrow (2 - \zeta);$ {\scriptsize{\tcp{Normalization}}} 
     
     $\{\Theta,\psi^{(t)}\} \leftarrow \text{Update model parameters using }\mathcal{L}_{stage1};$
     
  }

  \textcolor[HTML]{bc5a45}{\textit{$//$$\# Stage $ 2: Classifier Alignment$//$}}

    $\tilde{\mathcal{D}}^{(t)}\leftarrow \text{Expanded labelled data set using }\mathcal{D}_{l}^{(t)}, \mathcal{D}_{ul}^{(t)}, \gamma_{a}^{(t)};$

    $\tilde{\mu}_{k}^{(t)}, \tilde{\Sigma}_{k}^{(t)} \leftarrow \text{Estimate mean and variance using }\tilde{\mathcal{D}}^{(t)};$ {\scriptsize{\tcp{$\text{where } k\in{1,2,.,|\mathcal{C}^{(t)}|} $}}}
    % $\mu_{t}, \Sigma_{t} \leftarrow \mathcal{D}_{l}^{(t)}  \cup \tilde{\mathcal{D}}_{ul}^{(t)} $\\
    % $\tilde{\mu_{t}}, \tilde{\Sigma_{t}} \leftarrow [(\mu_{1}, \Sigma_{1}), (\mu_{2}, \Sigma_{2}) \dots (\mu_{t}, \Sigma_{t})]$\\
    \For{$e_{s2} \leftarrow 1$ to $E_{s2}$}{
     $\mathcal{L}_{stage2} \leftarrow \mathcal{L}_{ca}(\tilde{\mu}_{k}^{(1:t)}, \tilde{\Sigma}_{k}^{(1:t)});$ {\scriptsize{\tcp{$\text{Alignment loss for classifiers (Eq.~\ref{eq_l_ca})}$}}}

     $\psi^{(1:t)} \leftarrow \text{Update classifier parameters using }\mathcal{L}_{stage2}$;
     % Update $\psi^{(1:t)}$ using $\mathcal{L}_{ca}$\\
    
  }
}

\label{alg:tacle}
\end{algorithm}
% **************************************

% **************************************
\section{Experiments}
\label{sec:experiments}
Here, we discuss the datasets used, implementation details and experimental results of the proposed methodology.
\subsection{Datasets}
\label{subsec:datasets}
We evaluate our approach on three widely used SS-CIL datasets, which we briefly describe below. \\
\textbf{(i) CIFAR10~\cite{krizhevsky2009learning_cifar}:} This dataset comprises $32\times32$ images, with a total of 50,000 training images and 10,000 validation images distributed across 10 classes. Following the SS-CIL protocol, we structured the learning into 5 tasks, each involving the incremental learning of 2 classes (2-2-...-2) per task. In each task, the model has access to both labeled and unlabeled data. \\
\textbf{(ii) CIFAR100~\cite{krizhevsky2009learning_cifar}:} With a total of 100 classes, each consisting of $32 \times 32$ images, CIFAR100 presents 500 training and 100 validation images for each class. The SS-CIL protocol here spans 10 tasks, introducing 10 new classes in each task (10-10-...-10). In both CIFAR10 and CIFAR100 images are resized to $224 \times 224$ for compatibility with the pre-trained models.\\
\textbf{(iii) ImageNet-Subset100~\cite{tian2020contrastive_imagenet100}:} This dataset is a subset of ImageNet-1k~\cite{russakovsky2015imagenet}, containing 100 classes. All images were resized to $256\times256$ pixels and randomly cropped to $224\times224$ pixels during training. For each of the 100 classes, there are 1,300 training images and 50 testing images. 
For the SS-CIL protocol, it is structured into 20 tasks, introducing 5 new classes in each task (5-5-...-5). \\
\textbf{Supervision levels:} We evaluated our approach under different levels of supervision for labeled data proposed in NNCSL~\cite{kang2023soft_nncil_sscil3}. The percentage of labeled data used per task is set to $0.8\%, 5\%, \text{and } 25\%$ for CIFAR10 and CIFAR100. For ImageNet-Subset100, we use supervision levels of $1\%, 5\%, \text{and } 25\%$.

\subsection{Implementation details and Evaluation Protocol}
Inspired by SLCA~\cite{zhang2023slca}, we adopted a pre-trained ViT-B/16 backbone model for all our experiments. To test the approach's effectiveness for different pre-trained models, we have experimented with a supervised pre-trained model (trained on ImageNet-21k~\cite{ridnik2021imagenet_21k}) and also a self-supervised pre-trained model (trained on MoCo v3~\cite{chenempirical_mocov3}). 
Results on both CIFAR10 and CIFAR100 are reported for these pre-trained models, while for ImageNet-Subset100, we use the MoCo v3 pre-trained model (since the other model cannot be used due to data overlap). 
%Each task involved a two-stage training process: (i) 
In stage 1 of training, which focuses on feature representation learning, the model is trained for 10 epochs for CIFAR-10 and CIFAR-100 and for 5 epochs for ImageNet-Subset100.
We used an SGD optimizer with learning rate $0.005$, momentum 0.9 and weight decay of $5e^{-3}$ for all the experiments across all datasets. Batch sizes are set to 128 for CIFAR10 and CIFAR100 and 64 for ImageNet-Subset100. 
During stage 2 of training, the classifier alignment is performed for 5 epochs for all datasets. All the experiments are conducted on a system equipped with two NVIDIA RTX A5000 GPUs, each with 24GB of memory. We use the PyTorch deep learning library for our implementation.\\
\textbf{Evaluation protocol:} For fair comparison, our approach is evaluated using the same data splits and evaluation protocol proposed in NNCSL~\cite{kang2023soft_nncil_sscil3}. The evaluation considered both top-1 cumulative and average accuracy to assess the models. These metrics are calculated as follows, let $t$ represent the task ID, where $t\in{1,...,\mathcal{T}}$. 
Then $Acc_{1:t}^{t}$ denotes the model's accuracy on the test data of all tasks from $1$ to $t$ after learning task $t$. 
Consequently, upon completion of task $\mathcal{T}$, the average incremental accuracy is computed as $\frac{1}{\mathcal{T}}\sum_{t=1}^{\mathcal{T}} Acc_{1:t}^{t}$ and top1-cumulative accuracy is reported as $Acc_{1:\mathcal{T}}^{\mathcal{T}}$.

\setlength{\tabcolsep}{7pt}
\begin{table}[t]
  \caption{Average incremental accuracy on CIFAR10 after 5 tasks and CIFAR100 after 10 tasks for SS-CIL. The number in brackets indicates the number of exemplars; our approach does not use any exemplars. Here, $\star$: models trained from scratch, $\dag$: models initialized with MoCo v3 pretrained weights, and $\ddag$: models initialized with ImageNet pretrained weights; RN18: ResNet18 architecture, FT: fixed threshold.}
  % \label{tab:expts_on_cifar10_cifar100}
  \label{table_cifar_10_100}
  \centering
  \begin{adjustbox}{max width=\linewidth}
  \begin{tabular}{lcllllll}
    \hline\hline
    % Method & CIFAR-10 & CIFAR-100
    \textbf{\multirow{2}{*}{Method} }& \multirow{2}{*}{\textbf{Model}} & \multicolumn{3}{|c}{\rule{-2pt}{10pt} \textbf{CIFAR 100}} & \multicolumn{3}{|c}{\textbf{ CIFAR 10}} \\ \cline{3-8} 
    & &\multicolumn{1}{|c}{\rule{-2pt}{10pt} \textbf{$0.8\%$}} & \multicolumn{1}{c}{\textbf{$5\%$} }& \multicolumn{1}{c}{\textbf{$25\%$}} & \multicolumn{1}{|c}{\textbf{$0.8\%$}} & \multicolumn{1}{c}{\textbf{$5\%$}} & \multicolumn{1}{c}{\textbf{$25\% $} }\\
    \hline\hline
    Fine Tuning &  \multirow{2}{*}{RN18$^{\star}$} & \multicolumn{1}{|c}{1.8  $\pm$ 0.2} & \multicolumn{1}{c}{5.0  $\pm$ 0.3 }& \multicolumn{1}{c}{7.8  $\pm$ 0.1} & \multicolumn{1}{|c}{13.6 $\pm$ 2.9} & \multicolumn{1}{c}{18.2 $\pm$ 0.4} & \multicolumn{1}{c}{19.2 $\pm$ 2.2 } \\
    oEWC~\cite{kirkpatrick2017overcoming_reg_3}  &        & \multicolumn{1}{|c}{1.4  $\pm$ 0.1} & \multicolumn{1}{c}{4.7  $\pm$ 0.1} & \multicolumn{1}{c}{7.8  $\pm$ 0.4} & \multicolumn{1}{|c}{13.7 $\pm$ 1.2} & \multicolumn{1}{c}{17.6 $\pm$ 1.2} & \multicolumn{1}{c}{19.1 $\pm$ 0.8}  \\
    \hline
    ER~\cite{rolnick2019experience} (500)  &  \multirow{4}{*}{RN18$^{\star}$}  & \multicolumn{1}{|c}{8.2  $\pm$ 0.1} & \multicolumn{1}{c}{13.7 $\pm$ 0.6} & \multicolumn{1}{c}{17.1 $\pm$ 0.7}  & \multicolumn{1}{|c}{36.3 $\pm$ 1.1} & \multicolumn{1}{c}{51.9 $\pm$ 4.5} & \multicolumn{1}{c}{60.9 $\pm$ 5.7}  \\
    iCaRL~\cite{rebuffi2017icarl_algo_logit_1} (500)  & & \multicolumn{1}{|c}{3.6  $\pm$ 0.1} & \multicolumn{1}{c}{11.3 $\pm$ 0.3} & \multicolumn{1}{c}{27.6 $\pm$ 0.4}  & \multicolumn{1}{|c}{24.7 $\pm$ 2.3} & \multicolumn{1}{c}{35.8 $\pm$ 3.2} & \multicolumn{1}{c}{51.4 $\pm$ 8.4} \\
    FOSTER~\cite{wang2022foster} (500)  && \multicolumn{1}{|c}{4.7  $\pm$ 0.6} & \multicolumn{1}{c}{14.1 $\pm$ 0.6} & \multicolumn{1}{c}{21.7 $\pm$ 0.7} & \multicolumn{1}{|c}{43.3 $\pm$ 0.7} & \multicolumn{1}{c}{51.9 $\pm$ 1.3} & \multicolumn{1}{c}{57.1 $\pm$ 2.0}  \\
    X-DER~\cite{boschini2022class_xder} (500)   && \multicolumn{1}{|c}{8.9  $\pm$ 0.3} & \multicolumn{1}{c}{18.3 $\pm$ 0.5} & \multicolumn{1}{c}{23.9 $\pm$ 0.7} & \multicolumn{1}{|c}{33.4 $\pm$ 1.2} & \multicolumn{1}{c}{48.2 $\pm$ 1.7} & \multicolumn{1}{c}{58.9 $\pm$ 1.5}  \\
    \hline
    PseudoER~\cite{kang2023soft_nncil_sscil3} (500) & \multirow{4}{*}{RN18$^{\star}$}  & \multicolumn{1}{|c}{8.7  $\pm$ 0.4} & \multicolumn{1}{c}{11.4 $\pm$ 0.5} & \multicolumn{1}{c}{18.3 $\pm$ 0.2}& \multicolumn{1}{|c}{50.5 $\pm$ 0.1 }& \multicolumn{1}{c}{56.5 $\pm$ 0.6} & \multicolumn{1}{c}{57.0 $\pm$ 0.6}  \\
    CCIC~\cite{boschini2022continual_ccic_sscil2} (500)     && \multicolumn{1}{|c}{11.5 $\pm$ 0.7} & \multicolumn{1}{c}{19.5 $\pm$ 0.2} & \multicolumn{1}{c}{20.3 $\pm$ 0.3}& \multicolumn{1}{|c}{54.0 $\pm$ 0.2} & \multicolumn{1}{c}{63.3 $\pm$ 1.9} & \multicolumn{1}{c}{63.9 $\pm$ 2.6}  \\
    PAWS~\cite{assran2021semi_paws} (500)     && \multicolumn{1}{|c}{16.1 $\pm$ 0.4} & \multicolumn{1}{c}{21.2 $\pm$ 0.4} & \multicolumn{1}{c}{19.2 $\pm$ 0.4 }& \multicolumn{1}{|c}{51.8 $\pm$ 1.6} & \multicolumn{1}{c}{64.6 $\pm$ 0.6} & \multicolumn{1}{c}{65.9 $\pm$ 0.3} \\
    CSL~\cite{kang2023soft_nncil_sscil3}  (500)     && \multicolumn{1}{|c}{23.6 $\pm$ 0.3} & \multicolumn{1}{c}{26.2 $\pm$ 0.5} & \multicolumn{1}{c}{29.3 $\pm$ 0.3}& \multicolumn{1}{|c}{64.5 $\pm$ 0.7} & \multicolumn{1}{c}{69.6 $\pm$ 0.5} & \multicolumn{1}{c}{70.0 $\pm$ 0.4}  \\
    NNCSL~\cite{kang2023soft_nncil_sscil3} (500)     && \multicolumn{1}{|c}{27.4 $\pm$ 0.5} & \multicolumn{1}{c}{31.4 $\pm$ 0.4} & \multicolumn{1}{c}{35.3 $\pm$ 0.3}& \multicolumn{1}{|c}{73.2 $\pm$ 0.1} & \multicolumn{1}{c}{77.2 $\pm$ 0.2} & \multicolumn{1}{c}{77.3 $\pm$ 0.1}  \\
    \hline
    PseudoER~\cite{kang2023soft_nncil_sscil3} (5120)& \multirow{2}{*}{RN18$^{\star}$}  &  \multicolumn{1}{|c}{15.1 $\pm$ 0.2} & \multicolumn{1}{c}{24.9 $\pm$ 0.5} & \multicolumn{1}{c}{30.1 $\pm$ 0.7}&\multicolumn{1}{|c}{55.4 $\pm$ 0.5} & \multicolumn{1}{c}{70.0 $\pm$ 0.3} & \multicolumn{1}{c}{71.5 $\pm$ 0.2}  \\
    CICC~\cite{boschini2022continual_ccic_sscil2} (5120)    && \multicolumn{1}{|c}{12.0 $\pm$ 0.3} & \multicolumn{1}{c}{29.5 $\pm$ 0.4} & \multicolumn{1}{c}{44.3 $\pm$ 0.1}& \multicolumn{1}{|c}{55.2 $\pm$ 1.4} & \multicolumn{1}{c}{74.3 $\pm$ 1.7} & \multicolumn{1}{c}{84.7 $\pm$ 0.9}  \\
    ORDisCo~\cite{wang2021ordisco_sscil1} (12500)  &&        \multicolumn{1}{|c}{-}       &  \multicolumn{1}{c}{ -}      &        \multicolumn{1}{c}{ - }& \multicolumn{1}{|c}{41.7 $\pm$ 1.2} & \multicolumn{1}{c}{59.9 $\pm$ 1.4} & \multicolumn{1}{c}{67.6 $\pm$ 1.8}      \\
        CSL~\cite{kang2023soft_nncil_sscil3} (5120)      && \multicolumn{1}{|c}{23.7 $\pm$ 0.5} & \multicolumn{1}{c}{41.8 $\pm$ 0.4} & \multicolumn{1}{c}{50.3 $\pm$ 0.8}& \multicolumn{1}{|c}{64.3 $\pm$ 0.7} & \multicolumn{1}{c}{73.1 $\pm$ 0.3} & \multicolumn{1}{c}{73.9 $\pm$ 0.1}  \\
    NNCSL~\cite{kang2023soft_nncil_sscil3} (5120)    && \multicolumn{1}{|c}{27.5 $\pm$ 0.7} & \multicolumn{1}{c}{46.0 $\pm$ 0.2} & \multicolumn{1}{c}{56.4 $\pm$ 0.5}& \multicolumn{1}{|c}{73.7 $\pm$ 0.4} & \multicolumn{1}{c}{79.3 $\pm$ 0.3} & \multicolumn{1}{c}{81.0 $\pm$ 0.2}  \\
    \hline
    
    SLCA~\cite{zhang2023slca} (0)   &\multirow{3}{*}{ViTs$^{\dag}$}  & \multicolumn{1}{|c}{66.43 $\pm$ 0.04}         & \multicolumn{1}{c}{81.86 $\pm$ 0.02}         &\multicolumn{1}{c}{86.95 $\pm$ 0.01}& \multicolumn{1}{|c}{93.55 $ \pm$ 0.03}        &  \multicolumn{1}{c}{94.45 $\pm$ 0.01}        &\multicolumn{1}{c}{\textbf{96.19} $\pm$ 0.01}        \\
    
    SLCA+FT (0)  && \multicolumn{1}{|c}{71.67 $\pm$ 0.09}         & \multicolumn{1}{c}{83.96 $\pm$ 0.06}         & \multicolumn{1}{c}{86.91 $\pm$ 0.02}& \multicolumn{1}{|c}{94.07 $\pm$ 0.07}          & \multicolumn{1}{c}{95.35 $\pm$ 0.05}         & \multicolumn{1}{c}{96.08 $\pm$ 0.02}       \\
    
   \rowcolor[HTML]{EDDBC7} TACLE (ours) (0) & & \multicolumn{1}{|c}{\textbf{79.51} $\pm$ 0.08}         & \multicolumn{1}{c}{\textbf{85.58} $\pm$ 0.05}          & \multicolumn{1}{c}{\textbf{87.24} $\pm$ 0.02}& \multicolumn{1}{|c}{\textbf{94.59} $\pm$ 0.08}     & \multicolumn{1}{c}{\textbf{95.49} $\pm$ 0.05}         & \multicolumn{1}{c}{96.02 $\pm$ 0.01}       \\
    \hline
    
    SLCA~\cite{zhang2023slca} (0)         &\multirow{3}{*}{ViTs$^{\ddag}$}&  \multicolumn{1}{|c}{63.67 $\pm$ 0.03}         & \multicolumn{1}{c}{91.38 $\pm$ 0.02}         &\multicolumn{1}{c}{93.69 $\pm$ 0.01}&\multicolumn{1}{|c}{91.64 $\pm$ 0.02}          &  \multicolumn{1}{c}{97.79 $\pm$ 0.01}        & \multicolumn{1}{c}{98.56 $\pm$ 0.01}        \\
    
    SLCA+FT (0) & & \multicolumn{1}{|c}{88.23 $\pm$ 0.04}         & \multicolumn{1}{c}{93.30 $\pm$ 0.03}         & \multicolumn{1}{c}{94.08 $\pm 0.01$}& \multicolumn{1}{|c}{98.45 $\pm$ 0.03}     & \multicolumn{1}{c}{98.26 $\pm$ 0.02}          & \multicolumn{1}{c}{\textbf{98.89} $\pm$ 0.02}         \\
    
    \rowcolor[HTML]{EDDBC7} TACLE (ours) (0) & & \multicolumn{1}{|c}{\textbf{92.35} $\pm$ 0.06}         & \multicolumn{1}{c}{\textbf{93.59} $\pm$ 0.04}         & \multicolumn{1}{c}{\textbf{94.10} $\pm$ 0.02}& \multicolumn{1}{|c}{\textbf{98.61} $\pm$ 0.03}     & \multicolumn{1}{c}{\textbf{98.44} $\pm$ 0.03}           & \multicolumn{1}{c}{98.86 $\pm$ 0.02}         \\
    
\hline\hline
  \end{tabular}
  
  \end{adjustbox}
\end{table}

\subsection{Baselines:}
We compare our TACLE framework against both supervised CIL approaches and exemplar-based SS-CIL approaches.
In the realm of traditional approaches, we included online Elastic Weight Consolidation (oEWC)~\cite{kirkpatrick2017overcoming_reg_3}, a method that does not require replay buffers, making it a relevant comparison point for our buffer-free approach. For replay-based strategies, we compare with exemplar replay method~\cite{rolnick2019experience}, iCaRL~\cite{rebuffi2017icarl_algo_logit_1}, FOSTER,~\cite{wang2022foster} and XDER~\cite{boschini2022class_xder}. Additionally, we considere PseudoER~\cite{kang2023soft_nncil_sscil3}, a two-stage learning approach that combines Experience Replay (ER) with semi-supervised learning (PAWS)~\cite{assran2021semi_paws}. Among the SS-CIL approaches, we selected three major exemplar-based baselines: CCIC~\cite{boschini2022continual_ccic_sscil2}, ORDisCo~\cite{wang2021ordisco_sscil1}, and NNCSL~\cite{kang2023soft_nncil_sscil3}. Each of these approaches requires storing data from previous tasks in memory buffers. CCIC and NNCSL explicitly define their memory buffer sizes as either 500 or 5120, while ORDisCo stores all labeled data, resulting in a buffer size of 12500. In contrast, our proposed approach operates with a buffer size of 0, making it more realistic compared to these methods. 

Given a pre-trained ViT's backbone architecture, a direct comparison with previous approaches that utilized ResNet architectures may not be entirely fair. To address this, we included SLCA~\cite{zhang2023slca}, a powerful CIL technique that uses ViT pre-trained models from labeled data, and SLCA with fixed threhold $\gamma$ inspired by FixMatch~\cite{sohn2020fixmatch}, aiming to leverage unlabeled data. Both these methods serve as a baseline for an equitable comparison using ViT architectures, and we set the buffer size to 0 for consistency with our TACLE framework for these approaches.
% Given pre-trained ViT's backbone architectures, a direct comparison with previous approaches that utilized ResNet architectures may not be entirely fair. To address this, we included SLCA~\cite{zhang2023slca}, a powerful CIL technique that uses pre-trained models from labeled data. Additionally, we ran SLCA with fixed threhold $\gamma$ inspired by FixMatch~\cite{sohn2020fixmatch}, serve to leverage unlabeled data, which can be considered as a baseline using ViT architectures for a more equitable comparison. For both methods, we set the buffer size to 0 for consistency with our TACLE framework.

% \subsection{Experimental results}
\subsection{Experimental Results}
This section presents the experimental results achieved by our TACLE framework. Table~\ref{table_cifar_10_100} summarizes the comprehensive findings on the CIFAR10 and CIFAR100 datasets, considering different pre-trained models with varying labeled data percentages. The mean of average incremental accuracy over three seeds~\cite{kang2023soft_nncil_sscil3} are reported for a comprehensive evaluation. For the CIFAR10 dataset, in the challenging scenario where only $0.8\%$ labeled data is available, TACLE exhibits notable improvements over the baseline SLCA. In that context, when leveraging MoCo v3 as the pre-trained model, TACLE achieves a $1.04\%$ enhancement, while with ImageNet pre-training, it achieves a substantial $7\%$ improvement. As the percentage of labeled data increases to $5\%$, TACLE maintains its effectiveness, showcasing a $1.04\%$ improvement over SLCA with MoCo pre-training and a $0.65\%$ improvement with ImageNet pre-training. 

On CIFAR100 dataset, using the MoCo pre-trained model, TACLE achieves improvements of $13.08\%$ and $3.72\%$ over SLCA for $0.8\%$ and $5\%$ labeled data, respectively. The ImageNet pre-trained model also demonstrates significant gains, with improvements of $28.68\%$ and $2.21\%$ for $0.8\%$ and $5\%$ labeled data, respectively. As the percentage of labeled data increases, the contribution of TACLE or fixed threshold becomes less significant. In such cases, using even a small proportion of incorrectly pseudo-labeled unlabeled data may lead to a decrease in performance for pre-trained models. This trend is evident in the results for CIFAR-10 and CIFAR-100 with $25\%$ labeled data.

In ImageNet-Subset100, TACLE outperforms SLCA by $2.52\%$ and $3.13\%$ in the $1\%$ and $5\%$ labeled data settings, respectively. 
The improvements are prominent for the difficult scenarios, when the percentage of labeled data is less.
%Also, with an increased percentage of labeled data, pretrained models reach saturation, and improvements become insignificant. 
The complete results are given in Table~\ref{table_imagenet}.

\begin{table}[tb]
  % \caption{Average incremental accuracy of experimental results on ImageNet-Subset100 after 20 tasks for SS-CIL. The number in brackets indicates the buffer size for exemplars. Here, $\star$ denotes models trained from scratch and $\dag$ indicates models initialized with MoCo v3 pretrained weights.}
  \caption{Comparison of average incremental accuracy on ImageNet-Subset100 after 20 tasks for SS-CIL. The number in brackets: buffer size; $\star$: models trained from scratch and $\dag$: model initialized with MoCo v3 pretrained weights.}
  % \label{tab:expts_on_cifar10_cifar100}
  \label{table_imagenet}
  \centering
  \begin{adjustbox}{scale=0.7}
  \begin{tabular}{lclll}
    \hline\hline
    % Method & CIFAR-10 & CIFAR-100
    \multirow{2}{*}{Method} &\multirow{2}{*}{Model} &\multicolumn{3}{|c}{ImageNet100-Subset} \\ \cline{3-5} 
    & &\multicolumn{1}{|c}{\rule{-2pt}{10pt}$1\%$} & \multicolumn{1}{c}{$5\%$} & \multicolumn{1}{c}{$25\%$ }  \\
    \hline
    Fine-tuning    &\multirow{5}{*}{ResNet18$^{\star}$} &\multicolumn{1}{|c}{1.5  $\pm$ 0.2} &\multicolumn{1}{c}{ 2.7  $\pm$ 0.1} &\multicolumn{1}{c}{ 4.1  $\pm$ 0.2} \\
    ER~\cite{rolnick2019experience} (5120)      & &\multicolumn{1}{|c}{12.2 $\pm$ 0.8} & \multicolumn{1}{c}{26.3 $\pm$ 0.7} &\multicolumn{1}{c}{ 38.8 $\pm$ 1.0} \\
    FOSTER~\cite{wang2022foster} (5120)  & &\multicolumn{1}{|c}{14.8 $\pm$ 1.1} & \multicolumn{1}{c}{32.8 $\pm$ 0.7 }&\multicolumn{1}{c}{42.1 $\pm$ 1.5 }\\
    X-DER~\cite{boschini2022class_xder} (5120)   & &\multicolumn{1}{|c}{10.8 $\pm$ 1.1} & \multicolumn{1}{c}{27.4 $\pm$ 1.6} &\multicolumn{1}{c}{ 45.3 $\pm$ 1.0} \\
    CCIC~\cite{boschini2022continual_ccic_sscil2} (5120)    & &\multicolumn{1}{|c}{13.5 $\pm$ 1.2} & \multicolumn{1}{c}{19.5 $\pm$ 0.7 }& \multicolumn{1}{c}{25.9 $\pm$ 0.9 }\\
    \hline
    CSL~\cite{kang2023soft_nncil_sscil3} (5120)      &\multirow{2}{*}{ResNet18$^{\star}$} &\multicolumn{1}{|c}{26.8 $\pm$ 0.4} & \multicolumn{1}{c}{47.9 $\pm$ 0.2 }& \multicolumn{1}{c}{56.3 $\pm$ 0.5} \\
    NNCSL~\cite{kang2023soft_nncil_sscil3} (5120)    & &\multicolumn{1}{|c}{29.7 $\pm$ 0.4} & \multicolumn{1}{c}{51.3 $\pm$ 0.1} & \multicolumn{1}{c}{65.6 $\pm$ 0.3} \\
    \hline
    SLCA~\cite{zhang2023slca} (0) & \multirow{3}{*}{ViTs$^{\dag}$}&\multicolumn{1}{|c}{78.30 $\pm$ 0.04} & \multicolumn{1}{c}{79.29 $\pm$ 0.02} & \multicolumn{1}{c}{82.39 $\pm$ 0.01}\\
    SLCA+Fixed Threshold (0)& &\multicolumn{1}{|c}{79.72 $\pm$ 0.08} & \multicolumn{1}{c}{82.21 $\pm$ 0.05} & \multicolumn{1}{c}{\textbf{83.08} $\pm$ 0.02}\\
    
  \rowcolor[HTML]{EDDBC7}  TACLE (ours) (0) & &\multicolumn{1}{|c}{\textbf{80.82} $\pm$ 0.09} & \multicolumn{1}{c}{\textbf{82.42} $\pm$ 0.04} & \multicolumn{1}{c}{83.01 $\pm$ 0.02}\\
\hline \hline
  \end{tabular}
  
  \end{adjustbox}
\end{table}
%=======================================================IMAGENET 100 ===================================================================

% % **************************************
\section{Analysis and Ablation Studies}
\label{sec:ablation_studies}
\begin{figure}[tb!]
% \vspace{-1.0cm}
\centering
\input{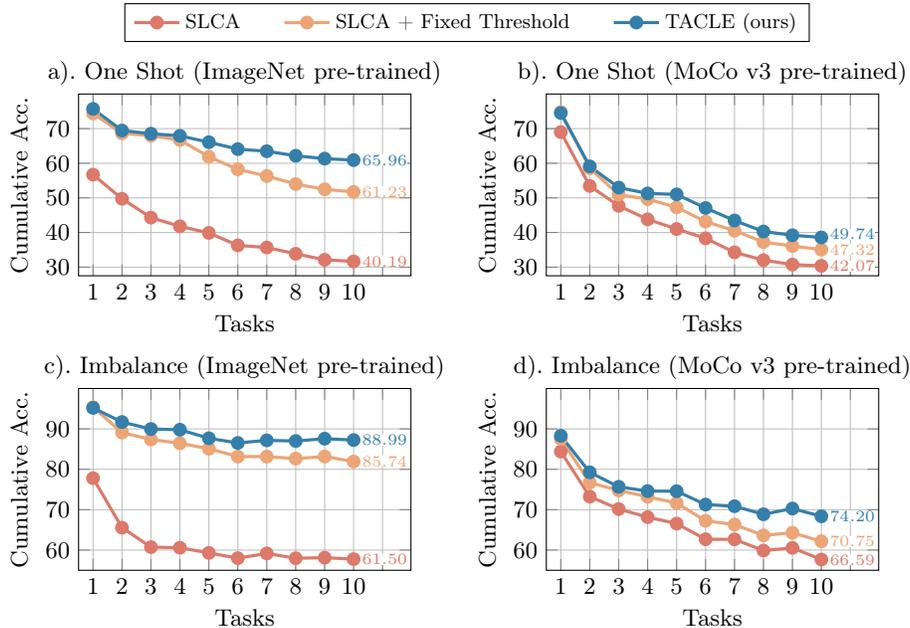}
\vspace{-0.3cm}
\caption{Analysis of one-shot SS-CIL and imbalance SS-CIL experiments. Experiments were conducted on CIFAR100 ($0.8\%$ labeled data for imbalance scenerio) with 10 tasks, reporting top-1 cumulative accuracy at the end of each task and average cumulative accuracy at the end of each plot. Results are presented for both pre-trained models.}
\vspace{-0.3cm}
\label{fig:one_shot_imbalance_ablation}
\end{figure}

\subsubsection{TACLE in Challenging Scenarios}
To evaluate the efficacy of the proposed TACLE framework in extreme scenarios, we conducted experiments under two challenging SS-CIL scenarios: one-shot EFSS-CIL and imbalanced EFSS-CIL.\\
\textbf{(i). One-shot EFSS-CIL:} In this scenario, each class has only one labeled sample, while the remaining data remains unlabeled. Experiments are carried out on CIFAR100 data with a 10-task configuration for one-shot EFSS-CIL. Fig.~\ref{fig:one_shot_imbalance_ablation}a illustrates the task-wise cumulative accuracy and average incremental accuracy for one-shot EFSS-CIL with ImageNet as the pre-trained model, and Fig.~\ref{fig:one_shot_imbalance_ablation}b shows the results with the MoCo pre-trained model. In both scenarios, the TACLE framework demonstrates significant improvements of $25.77\%$ (ImageNet pre-trained) and $7.67\%$ (MoCo v3 pre-trained) over the baseline SLCA. Table~\ref{table:one_shot_imagenet} shows the one-shot EFSS-CIL results on ImageNet-Subset100.\\
\textbf{(ii). Imbalance SS-CIL:} In this setup, the distribution of unlabeled data is imbalanced, deviating from traditional SS-CIL where unlabeled data is balanced. We introduced standard imbalance in unlabeled data with an imbalance ratio between minimum to maximum number of samples is 0.01 (This results in the minority class having 5 samples and the majority class having 500 samples). We considered $0.8\%$ labeled data on CIFAR100 data with a 10-task learning setup. Figure~\ref{fig:one_shot_imbalance_ablation}c and Figure~\ref{fig:one_shot_imbalance_ablation}d present the experimental results in these imbalance SS-CIL settings. These outcomes showcase the effectiveness of the TACLE framework in handling extreme EFSS-CIL scenarios.

% In this setup, the distribution of unlabeled data is imbalanced, which deviates from the traditional SS-CIL where unlabeled data is balanced. We created standard imbalance in unlabled data where the imbalance ratio of 0.01 ($\#$ of samples in class have least samples/ $\#$ of samples of class have maximum samples) and for labeled data we considered  $0.8\%$ labeled data on CIFAR100 data with 10 task learning. It will be like for unlable data minority class will have 5 samples and major class have 500 samples in the task. Figure~\ref{fig:one_shot_imbalance_ablation}.c and Figure~\ref{fig:one_shot_imbalance_ablation}.d present the experimental results on these challenging setting,  These outcomes showcase the effectiveness of the proposed components in TACLE framework to handling Imbalance SS-CIL scenarios.

\subsubsection{Ablation Study}
In this section, we provide a detailed study of the proposed components in the TACLE framework. Table~\ref{table_ablation} presents the detailed experimental results on the CIFAR100 dataset with $0.8\%$ labeled data using different pre-trained models. The baseline SLCA~\cite{zhang2023slca} utilizes only labeled data for stage 1 and stage 2 classifier alignment. SLCA + Fixed Threshold utilizes unlabeled data for training. Table~\ref{table_ablation} shows that incorporating each proposed component of TACLE has indeed improved the performance. Table~\ref{table:table_hyper_parameters} shows the impact of hyper-parameters ($\alpha,\:\beta$) in task-wise adaptive threshold (Eq.~\ref{eq_task_adap_th}). 
We observe that the results vary gracefully with change in these parameters. \\ 
\textbf{Discussion on limitations and future work:}
While TACLE excels in leveraging unlabeled data from the current task, it inherently assumes (as in the SS-CIL protocol) that the unlabeled data comes solely from the current task, whereas real-world scenarios may involve mixed data sources, including samples from previous tasks or outliers. 
Exploring these challenging and more realistic settings will be one of our future directions. 
Additionally, this framework can be extended to other tasks like object detection or segmentation.
% In this subsection, we provide a detailed study of the proposed components in the TaLE framework. Table~\ref{table_ablation} presents the detailed experimental results on the CIFAR100 dataset with $0.8\%$ labeled data using different pre-trained models. The baseline SLCA~\cite{zhang2023slca} utilizes only labeled data for stage 1 and stage 2 classifier alignment. SLCA + Fixed Threshold utilizes unlabeled data to improve the model learning performance. By incorporating proposed components C1: task-wise dynamic threshold (Eq. \ref{eq_lus}), C2: class-aware weighted cross-entropy loss (Eq. \ref{eq_stage1}), and C3: exploiting unlabeled data in classifier alignment (Eq. \ref{eq_l_ca}), we demonstrate the importance of each proposed component.
% The importance of each component in TaLE in the table 3, where in labeled data uses only pre-trained model to tune on this labeled data and to classfier alignment for also using labeled data. then we fixed threshold unsupervised loss is introduced, then the dynmaic threshold is introduced in the unsupervised loss. Then class aware weighted loss is introduced for both supervised and unsupervised loss. Then exploiting unlabeled data with along with above losses is our TaLE, which improves the further more.

% \begin{table}[t]

\begin{table}[t]
% \vspace{-1.0cm}
\centering
%===================================================================Ablation Table=============================================================
% \usepackage{pifont}
\newcommand{\cmark}{\ding{51}}%
\newcommand{\xmark}{\ding{55}}%

\caption{Ablation study on CIFAR100 dataset with ${0.8\%}$ labeled data. The average incremental accuracy is reported at the end of 10 tasks. The proposed components are denoted as C1: task-wise dynamic threshold (Eq. \ref{eq_lus}), C2: class-aware CE loss (Eq. \ref{eq_stage1}), C3: exploiting unlabeled data in stage 2 (Eq. \ref{eq_l_ca}).}
\centering
\label{table_ablation}
\begin{adjustbox}{scale=0.7}
\begin{tabular}{llllllll}
\toprule
\multirow{2}{*} {\textbf{Method} } & \multicolumn{2}{|c}{\textbf{Data}}  & \multicolumn{3}{|c}  {\textbf{Components} }  & \multicolumn{2}{|c}{\textbf{Pre-trained}} \\ 

\cline{2-8} 

 &\multicolumn{1}{|c}{\rule{-2pt}{10pt} Labelled} & \multicolumn{1}{c}{Unlabeled }& \multicolumn{1}{|c}{C1 }& \multicolumn{1}{c}{C2}  & \multicolumn{1}{c}{C3} & \multicolumn{1}{|c}{ImageNet}& \multicolumn{1}{c}{MoCo v3} \\

\midrule \rule{-2pt}{10pt} 

SLCA &\multicolumn{1}{|c}{\rule{-2pt}{10pt} \cmark} & \multicolumn{1}{c}{\xmark }& \multicolumn{1}{|c}{\xmark }& \multicolumn{1}{c}{\xmark }  & \multicolumn{1}{c}{\xmark }& \multicolumn{1}{|c}{63.37}& \multicolumn{1}{c}{66.43}  \\
SLCA + Fixed Threshold &\multicolumn{1}{|c}{\rule{-2pt}{10pt} \cmark} & \multicolumn{1}{c}{\cmark }& \multicolumn{1}{|c}{\xmark }& \multicolumn{1}{c}{\xmark }  & \multicolumn{1}{c}{\xmark }& \multicolumn{1}{|c}{88.23}& \multicolumn{1}{c}{71.67}  \\
\rowcolor[HTML]{EDDBC7} &\multicolumn{1}{|c}{\rule{-2pt}{10pt} \cmark} & \multicolumn{1}{c}{\cmark }& \multicolumn{1}{|c}{\cmark }& \multicolumn{1}{c}{\xmark }  & \multicolumn{1}{c}{\xmark }& \multicolumn{1}{|c}{89.10}& \multicolumn{1}{c}{75.29}  \\
\rowcolor[HTML]{EDDBC7}TACLE (ours) &\multicolumn{1}{|c}{\rule{-2pt}{10pt} \cmark} & \multicolumn{1}{c}{\cmark }& \multicolumn{1}{|c}{\cmark }& \multicolumn{1}{c}{\cmark }  & \multicolumn{1}{c}{\xmark }& \multicolumn{1}{|c}{91.32}& \multicolumn{1}{c}{77.19}  \\
\rowcolor[HTML]{EDDBC7}  &\multicolumn{1}{|c}{\rule{-2pt}{10pt} \cmark} & \multicolumn{1}{c}{\cmark }& \multicolumn{1}{|c}{\cmark }& \multicolumn{1}{c}{\cmark }  & \multicolumn{1}{c}{\cmark }& \multicolumn{1}{|c}{\textbf{92.35}}& \multicolumn{1}{c}{\textbf{79.51}}  \\
% fixed threshold $(\gamma)$&\multicolumn{1}{|c}{\rule{-2pt}{10pt} $\mathcal{L}_{s} + \mathcal{L}_{us}^{\gamma}$} & \multicolumn{1}{c}{$\mathcal{L}_{ca}(\mu_{k}, \Sigma_{k})$ }& \multicolumn{1}{|c}{88.23}& \multicolumn{1}{c}{71.67}   \\
 % task-wise dynamic threshold $(\gamma_{d}^{(t)})$ &\multicolumn{1}{|c}{\rule{-2pt}{10pt} $\mathcal{L}_{s} + \mathcal{L}_{us}^{\gamma_{d}^{(t)}}$} & \multicolumn{1}{c}{$\mathcal{L}_{ca}(\mu_{k}, \Sigma_{k})$ }& \multicolumn{1}{|c}{89.10}& \multicolumn{1}{c}{75.29}   \\
 % class aware weighted loss  &\multicolumn{1}{|c}{\rule{-2pt}{10pt} $\mathcal{L}_{s} \cdot w^{l} + \mathcal{L}_{us}^{\gamma_{d}^{(t)}} \cdot w^{ul}$} & \multicolumn{1}{c}{$\mathcal{L}_{ca}(\mu_{k}, \Sigma_{k})$ }& \multicolumn{1}{|c}{91.32}& \multicolumn{1}{c}{77.19}   \\
% exploiting unlabeled data in CA  &\multicolumn{1}{|c}{\rule{-2pt}{10pt} $\mathcal{L}_{s} \cdot w^{l} + \mathcal{L}_{us}^{\gamma_{d}^{(t)}} \cdot w^{ul}$} & \multicolumn{1}{c}{$\mathcal{L}_{ca}(\tilde{\mu}_{k}, \tilde{\Sigma}_{k})$ }& \multicolumn{1}{|c}{92.35}& \multicolumn{1}{c}{79.51}   \\ 

\bottomrule

\end{tabular}
\end{adjustbox}

% \vspace{-0.3cm}

% \label{fig:momp_vs_tasks}
\end{table}

\begin{table}[t]
    \vspace{0.5cm}

    \begin{minipage}{.4\textwidth}
        \centering 
        \captionof{table}{\smaller One-shot SS-CIL on ImageNet-Subset100 for 20 tasks}
            \scalebox{0.7}{
            \begin{tabular}{cc} % Add 5mm space after each column except the last
            \hline 
            \multicolumn{1}{c} {\textbf{Method}}  & \multicolumn{1}{|c}{\textbf{\textbf{Avg. inc. acc.}}} \\
        
            \hline 
            \multicolumn{1}{c}{ SLCA } & \multicolumn{1}{|c}{\rule{-2pt}{10pt} \textbf{$59.48$}} \\
            \multicolumn{1}{c}{ SLCA + Fixed Threshold } & \multicolumn{1}{|c}{\rule{-2pt}{10pt} \textbf{$61.32$}} \\
           \rowcolor[HTML]{EDDBC7}  \multicolumn{1}{c}{ TACLE (ours) } & \multicolumn{1}{|c}{\rule{-2pt}{10pt} \textbf{67.72}} \\
            
            \hline 
            \end{tabular}
            }
            \label{table:one_shot_imagenet}
    \end{minipage}
    \hfill
    \begin{minipage}{.56\textwidth}
    %\vspace{-0.45cm}
    \captionof{table}{\smaller Impact of threshold hyper-parameters $\alpha$ and $\beta$ on CIFAR100 dataset.}
    %\vspace{0.05cm}
        \centering 
        \scalebox{0.6}{
      \begin{tabular}{l|ccc}
  \hline
  \diagbox[width=1em, height=1em]{$\alpha  \downarrow$}{\text{\quad $\beta$  $\rightarrow$}} & \multicolumn{1}{|c}{\rule{-2pt}{10pt} \textbf{$0.6$}} & \multicolumn{1}{c}{\textbf{$0.65$} } & \multicolumn{1}{c}{\textbf{$0.7$} }\\
  \hline
 \multicolumn{1}{c}{ \textbf{$0.45$} } & \multicolumn{1}{|c}{\rule{-2pt}{10pt} \textbf{$92.12$}} & \multicolumn{1}{c}{\textbf{$91.78$} }& \multicolumn{1}{c}{\textbf{$92.07$} } \\
            \multicolumn{1}{c}{ \textbf{$0.50$} } & \multicolumn{1}{|c}{\rule{-2pt}{10pt} \textbf{$91.96$}} & \multicolumn{1}{c}{\textbf{92.35} }& \multicolumn{1}{c}{\textbf{$91.01$} } \\
            \multicolumn{1}{c}{ \textbf{$0.55$} } & \multicolumn{1}{|c}{\rule{-2pt}{10pt} \textbf{$91.86$}} & \multicolumn{1}{c}{\textbf{$91.01$} }& \multicolumn{1}{c}{\textbf{$90.52$} } \\
  \hline
\end{tabular}        }
        \label{table:table_hyper_parameters}
    \end{minipage}
\end{table}

\section{Conclusion}
This paper introduces the framework TACLE, an exemplar-free approach for SS-CIL. TACLE achieves state-of-the-art results on several benchmark datasets designed for SS-CIL by leveraging pre-trained models without exemplars. The proposed approach incorporates three key components to effectively utilize unlabeled data: (i) Task-wise adaptive threshold: facilitating effective utilization of unlabeled data, (ii) Class-aware weighted loss: improving performance on under-represented classes. (iii) Exploiting unlabeled data for classifier alignment. TACLE demonstrates its effectiveness not only under standard EFSS-CIL settings but also in extreme scenarios like one-shot EFSS-CIL and imbalanced EFSS-CIL. A comprehensive analysis conducted on various datasets underscores the significant improvements achieved by TACLE.\\

% ****************************************
\singleappendix{Appendix}
\label{sec:supplementary}

\subsection{Effect of hyper-parameters $\alpha \text{ and } \beta$ on task-wise threshold } 
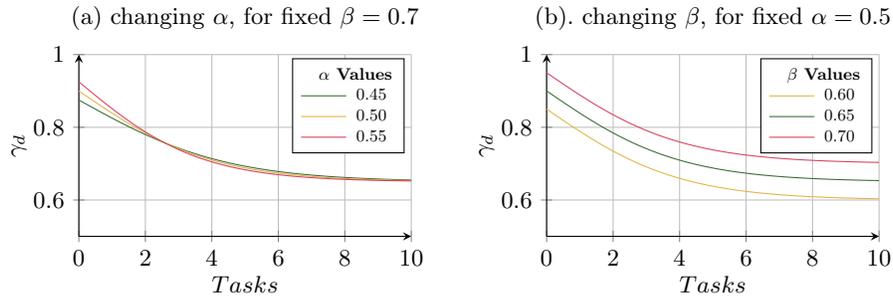
\begin{figure}[b]
% \vspace{-1.0cm}
\centering
\begin{tikzpicture}
    \definecolor{color0}{HTML}{E3B448} 
    \definecolor{color1}{HTML}{3A6B35}
    \definecolor{color2}{HTML}{D64161} 
    
 .  \begin{groupplot}[group style={group size=2 by 1, horizontal sep = 1.8cm, vertical sep = 1.8cm, },height=4cm,width=6cm, ]

    % alpha varies, beta fixed
        \nextgroupplot[title={(a) changing $\alpha$, for fixed $\beta=0.7$}, axis lines = left, xlabel =$Tasks$, ylabel = $\gamma_{d}$, grid=major,
        ymin=0.5, ymax=1,
        legend style={nodes={scale=0.7, transform shape}, mark options = {mark=*, mark size=2pt}},
        x label style={at={(axis description cs:0.5,0.05)},anchor=north},
        y label style={at={(axis description cs:0.05,0.5)},anchor=north},] 

        \addlegendimage{empty legend}
        \addlegendentry{\hspace{-.6cm}\textbf{$\alpha$ Values}}
        \addplot[domain=0:10, samples=100, color=color1, ]{(0.45/(1+exp(0.45*x))+0.65};
        \addlegendentry{$0.45$}
        \addplot[domain=0:10, samples=100, color=color0, ]{(0.5/(1+exp(0.5*x))+0.65};
        \addlegendentry{$0.50$}
        \addplot[domain=0:10, samples=100, color=color2, ]{(0.55/(1+exp(0.55*x))+0.65};
        \addlegendentry{$0.55$}
        
    % alpha fixed, beta varies
        \nextgroupplot[axis lines = left, xlabel =\textbf{$Tasks$}, ylabel = \textbf{$\gamma_{d}$}, title={(b). changing $\beta$, for fixed $\alpha = 0.5$}, grid=major,    
        ymin=0.5, ymax=1,
        legend style={nodes={scale=0.7, transform shape}, mark options = {mark=*, mark size=2pt}},
        x label style={at={(axis description cs:0.5,0.05)},anchor=north},
        y label style={at={(axis description cs:0.05,0.5)},anchor=north},] 
        % legend image post style={mark=*, mark size=2pt}]
        
        \addlegendimage{empty legend}
        \addlegendentry{\hspace{-.6cm}\textbf{$\beta$ Values}}
        \addplot[domain=0:10, samples=100, color=color0, ]{(0.5/(1+exp(0.5*x))+0.6};
        \addlegendentry{$0.60$}
        \addplot[domain=0:10, samples=100, color=color1, ]{(0.5/(1+exp(0.5*x))+0.65};
        \addlegendentry{$0.65$}
        \addplot[domain=0:10, samples=100, color=color2, ]{(0.5/(1+exp(0.5*x))+0.70};
        \addlegendentry{$0.70$}

    \end{groupplot}

\end{tikzpicture}
\caption{Task-wise adaptive threshold output values by changing hyper-paraneters ($\alpha, \beta$).}
\label{fig:variation_on_hyperparameters}
\end{figure}
This section analyzes the impact of hyper-parameters $\alpha \text{ and } \beta$ on the task-wise adaptive threshold defined by the equation:
\begin{equation}
\gamma_{a}^{(t)} = \frac{\alpha}{1+e^{\alpha t}} + \beta,
\label{eq_task_adap_th1}
\end{equation}
Figure~\ref{fig:variation_on_hyperparameters} illustrates the behavior of the task-wise adaptive threshold as we vary $\alpha \text{ and } \beta$ . Table~\ref{tab:hyper_parameter} shows the average incremental accuracy achieved on the CIFAR-100 dataset with $0.8\%$ labeled data per class across 10 incremental tasks.

As shown in Figure~\ref{fig:variation_on_hyperparameters}, the threshold value generally decreases with increasing task number (t). This aligns with the desired behavior of incorporating more unlabeled data as the number of labeled samples grows. The experiment results in Table~\ref{tab:hyper_parameter} suggest that the choice of $\alpha \text{ and } \beta$ impacts performance on incremental learning. For example, the configuration with $\alpha = 0.55 \text{ and } \beta=0.7$ leads to a lower average accuracy. This is likely due to a high threshold, which hinders the effective utilization of unlabeled data. We opted for this decaying threshold function inspired by the inverse sigmoid due to its simplicity and control over the initial and final threshold values. This allows for a smooth decrease in the threshold as tasks progress, enabling the model to leverage more unlabeled data effectively over time.

% Figure~\ref{fig:variation_on_hyperparameters} shows the behaviour of task-wise adaptive threshold by changing hyper-parameters ($\alpha, \beta$). Table~\ref{tab:hyper_paranmeter} shows the average incremental accuracy experimental results on the CIFAR100 dataset with $0.8\%$ label data scenario with 10 tasks. For values $\alpha = 0.55, \beta=0.7$, the experimental reulst reduced due to its hight threshold making less effectively utilization of unlabled data.

\setlength{\tabcolsep}{7pt}

\begin{table}[ht]
    \centering
    \captionof{table}{\smaller Impact of threshold hyper-parameters $\alpha$ and $\beta$ on CIFAR100 dataset.}
    \label{tab:hyper_parameter}
    %\vspace{0.05cm}
        \centering 
        
      \begin{tabular}{l|ccc}
  \hline
  \diagbox[width=1em, height=1em]{$\alpha  \downarrow$}{\text{\quad $\beta$  $\rightarrow$}} & \multicolumn{1}{|c}{\rule{-2pt}{10pt} \textbf{$0.60$}} & \multicolumn{1}{c}{\textbf{$0.65$} } & \multicolumn{1}{c}{\textbf{$0.70$} } \\
  \hline
 \multicolumn{1}{c}{ \textbf{$0.45$} } & \multicolumn{1}{|c}{\rule{-2pt}{10pt} \textbf{$92.12$}} & \multicolumn{1}{c}{\textbf{$91.78$} }& \multicolumn{1}{c}{\textbf{$92.07$} } \\
            \multicolumn{1}{c}{ \textbf{$0.50$} } & \multicolumn{1}{|c}{\rule{-2pt}{10pt} \textbf{$91.96$}} & \multicolumn{1}{c}{\textbf{92.35} }& \multicolumn{1}{c}{\textbf{$91.01$} } \\
            \multicolumn{1}{c}{ \textbf{$0.55$} } & \multicolumn{1}{|c}{\rule{-2pt}{10pt} \textbf{$91.86$}} & \multicolumn{1}{c}{\textbf{$91.01$} }& \multicolumn{1}{c}{\textbf{$90.52$} } \\
  \hline
\end{tabular}    

\end{table}

%========================================================================================================================
% \section{cifra100 results alpha-beta variations}
\subsection{Task-wise cumulative accuracy results}

% 1. plot cifar 0.8 = MoCo
% 2. plot cifar 5 = MoCo
% 3. plot cifar 0.8 = ImageNet
% 4. plot cifar 5 = ImageNet

% \color{red}Avg inc accuracy should match with tables reported in the paper.\color{black}

\begin{figure}[t]
% \vspace{-1.0cm}
\centering
\begin{tikzpicture}
    % \begin{groupplot}[group style={group size=2 by 2},height=4cm,width=8cm]
    \definecolor{color0}{HTML}{DC796A} %RED
    \definecolor{color1}{HTML}{ECA576} %Orange
    \definecolor{color2}{HTML}{357FAB} %BLUE
    
 .  \begin{groupplot}[group style={group size=2 by 2, horizontal sep = 1.8cm, vertical sep = 1.5cm, },height=3.8cm,width=5.6cm]
 
    \nextgroupplot[xlabel={Tasks}, ylabel={Cumulative Acc.}, 
                   x label style={at={(axis description cs:0.5,0.05)},anchor=north},
                   y label style={at={(axis description cs:0.05,0.5)},anchor=north},
                   xmin=0.5, xmax=11.98, xtick={1,2,3,4,5,6,7,8,9,10,11}, 
                   xticklabels={1,2,3,4,5,6,7,8,9,10}, 
                   ymin=50, ymax=100, ytick={50,60,70,80,90,100}, 
                   title={\smaller a). {0.8\% labeled (ImageNet pre-trained)}},
                   title style={anchor=north, yshift=2.5ex},
                   % legend pos=north west,
                   legend style={at={(2.2,1.5)},draw=black,legend columns=3, legend entries={SLCA, SLCA + Fixed Threshold, TACLE (ours)}, legend cell align={left},/tikz/every even column/.append style={column sep=0.5cm},nodes={scale=0.85, transform shape}},
                   grid=major]

        % [77.8, 64.3, 61.3, 61.42, 60.8, 61.05, 62.76, 62.04, 62.24, 63.08]: 63.67
        \addplot[very thick, color= color0, mark=*, mark options={solid}]  coordinates {
                        (1, 77.8)
                        (2, 64.3)
                        (3, 61.3)
                        (4, 61.42)
                        (5, 60.8)
                        (6, 61.05)
                        (7, 62.76)
                        (8, 62.04)
                        (9, 62.24)
                        (10, 63.08)
                    };
                   \node [right, fill=white,  inner sep=0pt] at (axis cs:10,63.08) {\color{color0} \fontsize{6}{6} \textbf{$63.67$}};

        % [96.3, 91.75, 90.13, 88.85, 87.64, 85.65, 85.81, 85.38, 85.68, 85.12]: 88.23           
        \addplot[very thick, color= color1, mark=*, mark options={solid}]  coordinates {
                        (1, 96.3)
                        (2, 91.75)
                        (3, 90.13)
                        (4, 88.85)
                        (5, 87.64)
                        (6, 85.65)
                        (7, 85.81)
                        (8, 85.38)
                        (9, 85.68)
                        (10, 85.12)
                    };
                
                    \node [right, fill=white,  inner sep=0pt] at (axis cs:10,85.12) {\color{color1} \fontsize{6}{6} \textbf{$88.23$}};
                    
        % [97.6, 95.7, 93.83, 93.08, 91.98, 90.78, 90.77, 90.06, 90.02, 89.63] : 92.35
        \addplot[very thick, color= color2, mark=*, mark options={solid}]  coordinates {
                        (1, 97.6)
                        (2, 95.7)
                        (3, 93.83)
                        (4, 93.08)
                        (5, 91.98)
                        (6, 90.78)
                        (7, 90.77)
                        (8, 90.06)
                        (9, 90.02)
                        (10, 89.63)
                    };
                   
                    \node [right, fill=white,  inner sep=0pt] at (axis cs:10,89.63) {\color{color2} \fontsize{6}{6} \textbf{$92.35$}};
        
    \nextgroupplot[xlabel={Tasks}, ylabel={Cumulative Acc.}, xmin=0.5, xmax=11.98, xtick={1,2,3,4,5,6,7,8,9,10,11},
    xticklabels={1,2,3,4,5,6,7,8,9,10}, 
                   x label style={at={(axis description cs:0.5,0.05)},anchor=north},
                   y label style={at={(axis description cs:0.05,0.5)},anchor=north},
                   ymin=50, ymax=100, ytick={50, 60,70,80,90,100}, 
                   title={\smaller b). {0.8\% labeled (MoCo v3 pre-trained)}},
                   title style={anchor=north, yshift=2.5ex},
                   % legend pos=north west,
                   grid=major,]
       
        % [84.4, 73.1, 69.97, 67.75, 66.24, 63.15, 62.03, 59.84, 60.06, 57.8]: 66.43
        \addplot[very thick, color= color0, mark=*, mark options={solid}]  coordinates {
                        (1, 84.4)
                        (2, 73.1)
                        (3, 69.97)
                        (4, 67.75)
                        (5, 66.24)
                        (6, 63.15)
                        (7, 62.03)
                        (8, 59.84)
                        (9, 60.06)
                        (10, 57.8)
                    };

                \node [right, fill=white,  inner sep=0pt] at (axis cs:10,57.8) {\color{color0} \fontsize{6}{6} \textbf{$66.43$}};

        % [88.3, 77.05, 74.9, 73.82, 72.24, 68.42, 67.63, 65.05, 65.86, 63.59]: 71.67
        \addplot[very thick, color= color1, mark=*, mark options={solid}]  coordinates {
                        (1, 88.3)
                        (2, 77.05)
                        (3, 74.9)
                        (4, 73.82)
                        (5, 72.24)
                        (6, 68.42)
                        (7, 67.63)
                        (8, 65.05)
                        (9, 65.86)
                        (10, 63.59)
                    };

                \node [right, fill=white,  inner sep=0pt] at (axis cs:10,63.59) {\color{color1} \fontsize{6}{6} \textbf{$71.67$}};

        % [89.2, 83.65, 81.07, 80.68, 79.74, 77.52, 76.53, 75.36, 76.5, 74.86]: 79.51
        \addplot[very thick, color= color2, mark=*, mark options={solid}]  coordinates {
                        (1, 89.2)
                        (2, 83.65)
                        (3, 81.07)
                        (4, 80.68)
                        (5, 79.74)
                        (6, 77.52)
                        (7, 76.53)
                        (8, 75.36)
                        (9, 76.5)
                        (10, 74.86)
                    };

                \node [right, fill=white,  inner sep=0pt] at (axis cs:10,74.86) {\color{color2} \fontsize{6}{6} \textbf{$79.51$}};

    \nextgroupplot[xlabel={Tasks}, ylabel={Cumulative Acc.}, xmin=0.5, xmax=11.98, xtick={1,2,3,4,5,6,7,8,9,10,11}, 
    xticklabels={1,2,3,4,5,6,7,8,9,10}, 
                   x label style={at={(axis description cs:0.5,0.05)},anchor=north},
                   y label style={at={(axis description cs:0.05,0.5)},anchor=north},
                   ymin=85, ymax=100, ytick={85,90,95,100},  
                   title={\smaller c). {5\% footnote (ImageNet pre-trained)}},
                   title style={anchor=north, yshift=2.5ex},
                   % legend pos=north west,
                   grid=major]

        %[97.5, 95.3, 93.4, 91.3, 90.56, 89.17, 89.37, 89.41, 89.43, 88.44]: 91.38
        \addplot[very thick, color= color0, mark=*, mark options={solid}]  coordinates {
                        (1, 97.5)
                        (2, 95.3)
                        (3, 93.4)
                        (4, 91.3)
                        (5, 90.56)
                        (6, 89.17)
                        (7, 89.37)
                        (8, 89.41)
                        (9, 89.43)
                        (10, 88.44)
                    };
                \node [right, fill=white,  inner sep=0pt] at (axis cs:10,88.44) {\color{color0} \fontsize{6}{6} \textbf{$91.38$}};
                
        % [99.1, 96.8, 95.27, 93.92, 92.86, 91.45, 91.2, 90.96, 91.22, 90.26]: 93.30
        \addplot[very thick, color= color1, mark=*, mark options={solid}]  coordinates {
                        (1, 99.1)
                        (2, 96.8)
                        (3, 95.27)
                        (4, 93.92)
                        (5, 92.86)
                        (6, 91.45)
                        (7, 91.2)
                        (8, 90.96)
                        (9, 91.22)
                        (10, 90.26)
                    };
                \node [right, fill=white,  inner sep=0pt] at (axis cs:10,90.26) {\color{color1} \fontsize{6}{6} \textbf{$93.30$}};

        % [99.3, 97.4, 95.8, 94.08, 93.34, 92.07, 91.74, 90.58, 91.1, 90.51]: 93.59
        \addplot[very thick, color= color2, mark=*, mark options={solid}]  coordinates {
                        (1, 99.3)
                        (2, 97.4)
                        (3, 95.8)
                        (4, 94.08)
                        (5, 93.34)
                        (6, 92.07)
                        (7, 91.74)
                        (8, 90.58)
                        (9, 91.1)
                        (10, 90.51)
                    };

                \node [right, fill=white,  inner sep=0pt] at (axis cs:10,92) {\color{color2} \fontsize{6}{6} \textbf{$93.59$}};

    \nextgroupplot[xlabel={Tasks}, ylabel={Cumulative Acc.}, xmin=0.5, xmax=11.98, xtick={1,2,3,4,5,6,7,8,9,10,11},
    xticklabels={1,2,3,4,5,6,7,8,9,10}, 
                   x label style={at={(axis description cs:0.5,0.05)},anchor=north},
                   y label style={at={(axis description cs:0.05,0.5)},anchor=north},
                   ymin=67.50, ymax=100, ytick={70, 80,90,100},  
                   title={\smaller d). {5\% labeled (MoCo v3 pre-trained)}},
                   title style={anchor=north, yshift=2.5ex},
                   % legend pos=north west,
                   grid=major]
       
        % [94.6, 88.15, 84.0, 83.15, 82.5, 79.27, 78.06, 77.19, 76.72, 74.96]: 81.86
        \addplot[very thick, color= color0, mark=*, mark options={solid}]  coordinates {
                        (1, 94.6)
                        (2, 88.15)
                        (3, 84.0)
                        (4, 83.15)
                        (5, 82.5)
                        (6, 79.27)
                        (7, 78.06)
                        (8, 77.19)
                        (9, 76.72)
                        (10, 74.96)
                    };

                \node [right, fill=white,  inner sep=0pt] at (axis cs:10, 74.96) {\color{color0} \fontsize{6}{6} \textbf{$81.86$}};

        % [95.1, 90.15, 86.0, 85.05, 83.68, 82.08, 80.76, 79.49, 79.27, 78.04]: 83.96
        \addplot[very thick, color= color1, mark=*, mark options={solid}]  coordinates {
                        (1, 95.1)
                        (2, 90.15)
                        (3, 86.0)
                        (4, 85.05)
                        (5, 83.68)
                        (6, 82.08)
                        (7, 80.76)
                        (8, 79.49)
                        (9, 79.27)
                        (10, 78.04)
                    };
                \node [right, fill=white,  inner sep=0pt] at (axis cs:10,78.04) {\color{color1} \fontsize{6}{6} \textbf{$83.96$}};

        % [95.7, 91.1, 87.5, 86.65, 85.68, 83.95, 82.61, 81.37, 81.13, 79.81]: 85.58
        \addplot[very thick, color= color2, mark=*, mark options={solid}]  coordinates {
                        (1, 95.7)
                        (2, 91.1)
                        (3, 87.5)
                        (4, 86.65)
                        (5, 85.68)
                        (6, 83.95)
                        (7, 82.61)
                        (8, 81.37)
                        (9, 81.13)
                        (10, 79.81)
                    };

                \node [right, fill=white,  inner sep=0pt] at (axis cs:10,79.81) {\color{color2} \fontsize{6}{6} \textbf{$85.58$}};
                
    \end{groupplot}
\end{tikzpicture}

\vspace{-0.3cm}
\caption{ Analysis for CIFAR100 datasets for different methods. Experiments
were conducted for 0.8\% and 5\% labeled data with 10 tasks,
reporting top-1 cumulative accuracy at the end of each task and average cumulative
accuracy at the end of each plot. Results are presented for both pre-trained models.}
\vspace{-0.3cm}
\label{fig:0.8_5_CIFAR}
\end{figure}
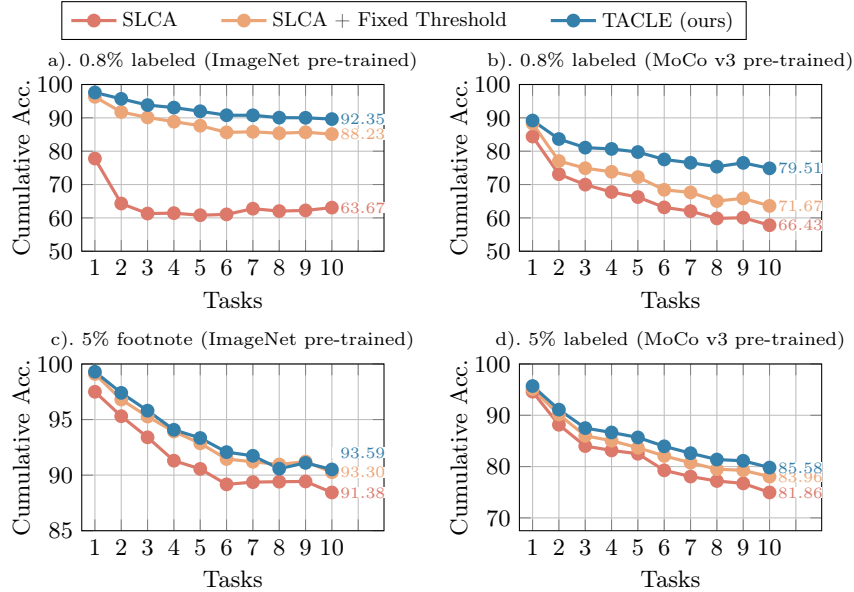

In this section, we report the task-wise cumulative accuracy results for the proposed approach TACLE, SLCA, and SLCA+Fixed threshold. Figure \ref{fig:0.8_5_CIFAR} presents the results for CIFAR100 with $0.8\%$ and $5\%$ labeled data settings for the EFSS-CIL protocol. We also report the average incremental accuracy at the end of the task for both cases where two different pre-trained models are used for model weight initialization. The proposed TACLE outperforms the baselines by a significant margin in the both the scenarios. 
% \color{red} Figure\ref{fig:0.8_5_CIFAR}c and Figure\ref{fig:0.8_5_CIFAR}d sclae are not same. If scale is same then its overlaps. Please check this.\color{black}

%========================================================================================================================

\newpage

\section{Visualization of features: SLCA vs TACLE (task 1,5,9)}
\begin{figure}[t!]
    \centering
    \begin{subfigure}{\textwidth}
        \centering
        \includegraphics[scale=0.40]{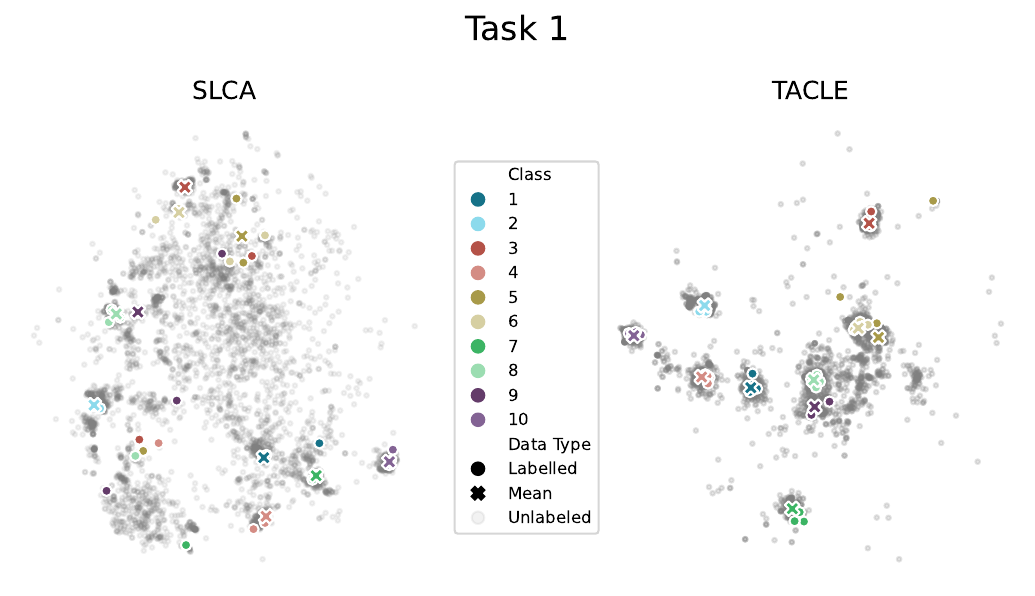}
        % \hfill
        \includegraphics[scale=0.40]{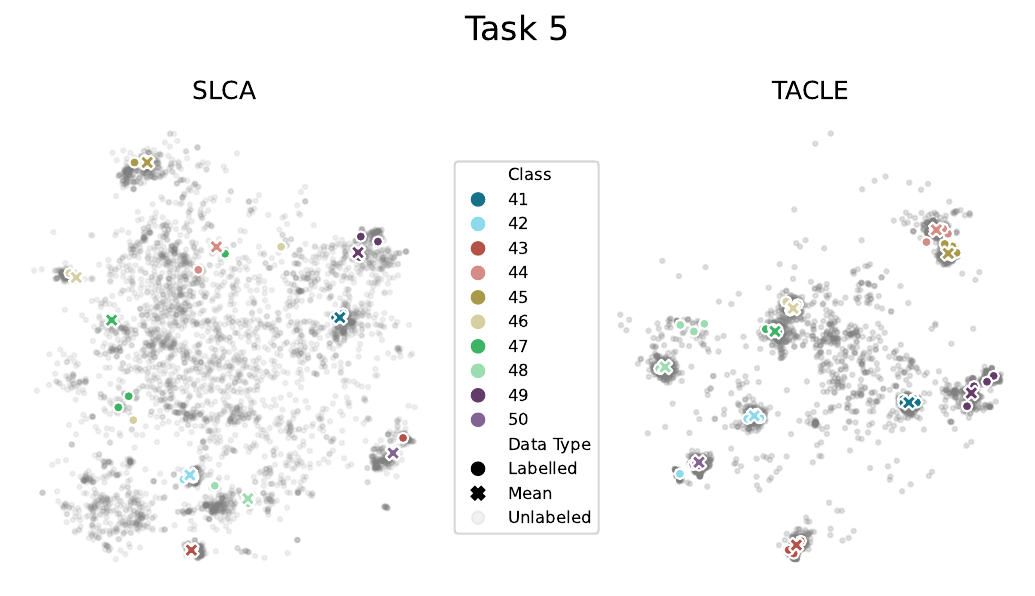}
        % \hfill
        \includegraphics[scale=0.40]{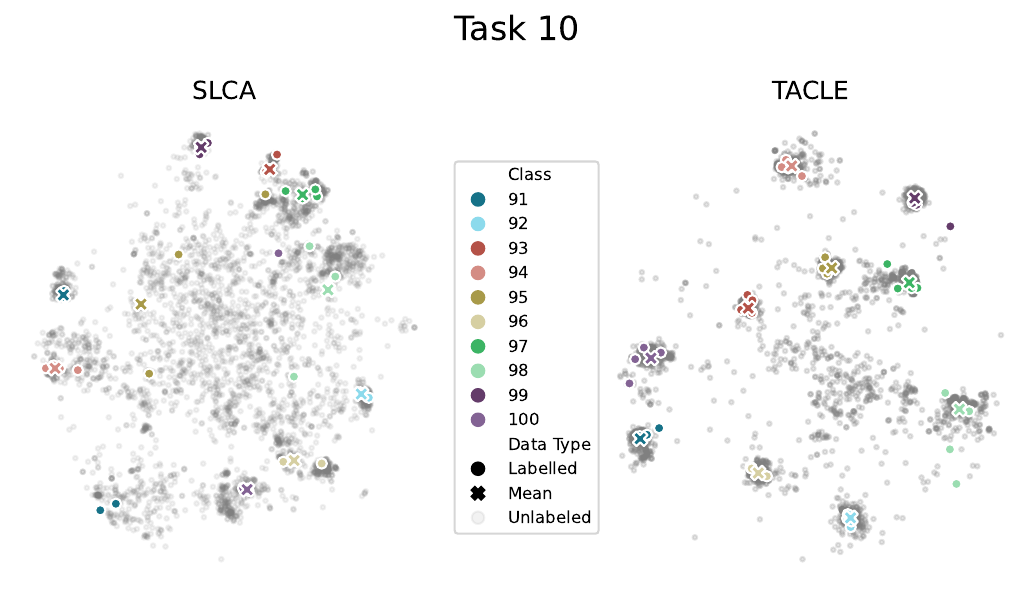}
    \end{subfigure}
    \caption{ t-SNE visualization of SLCA vs TACLE for given task id 1, 5, and 10. Each point represents image feature vector of dimension 768 (using ImageNet as pre-trained model). }
    \label{fig:tsne_imagenet}
\end{figure}
\begin{figure}[t!]
    \centering
    \begin{subfigure}{\textwidth}
        \centering
        \includegraphics[scale=0.4]{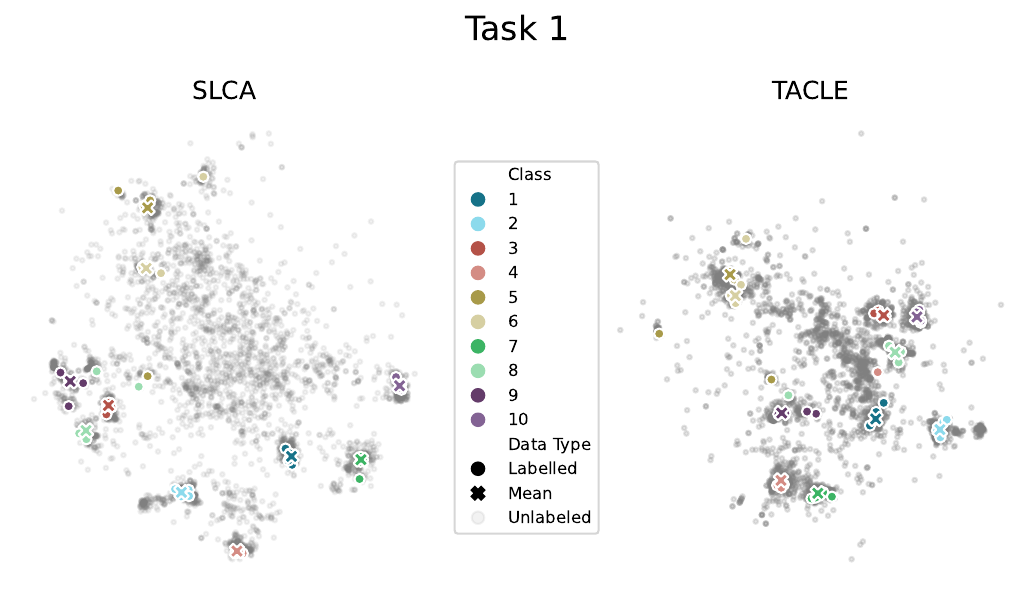}
        % \hfill
        \includegraphics[scale=0.4]{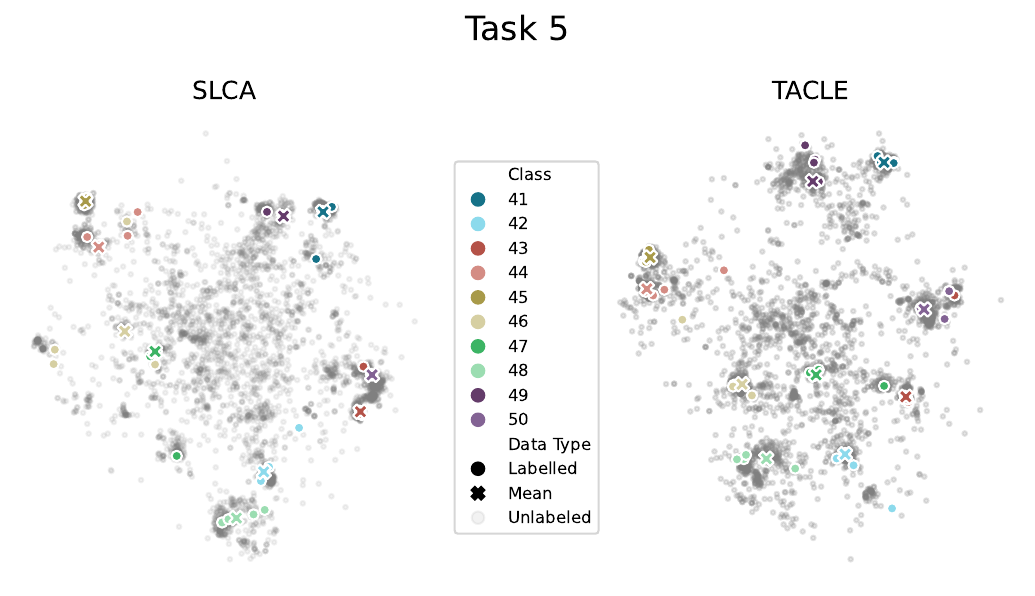}
        % \hfill
        \includegraphics[scale=0.4]{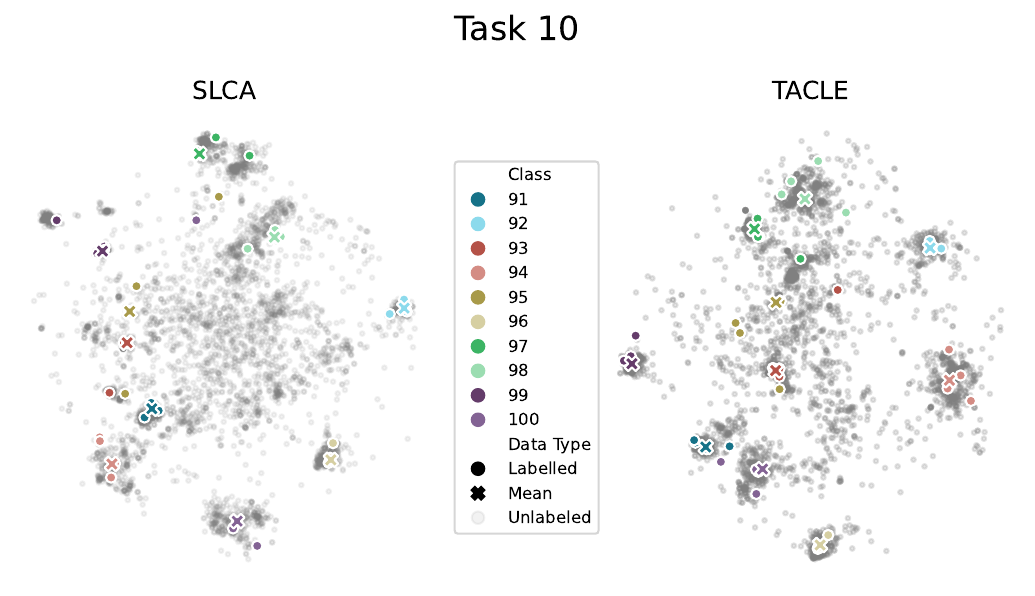}
    \end{subfigure}
     \caption{ t-SNE visualization of SLCA vs TACLE for given task id 1, 5, and 10. Each point represents image feature vector of dimension 768 (using moco V3 as pre-trained model). }
    \label{fig:tsne_moco}
\end{figure}

To visualize the clustering of unlabeled and labeled data, we employ t-SNE dimensionality reduction on the image features extracted from the model feature extractor ($\Theta$), which shares parameters across all tasks. We consider 4 labeled data points from each class, one class prototype for each, and all the task's unlabeled data (this is the data samples in CIFAR100 with $0.8\%$ at every task). Figures~\ref{fig:tsne_imagenet} and~\ref{fig:tsne_moco} depict t-SNE plots for both the SLCA approach (which utilizes only labeled data) and our TACLE framework after tasks 1, 5, and 10. These plots consider two pre-trained models for initial model weight initialization: ImageNet and MoCo v3. We observe that, by leveraging unlabeled data, proposed TACLE achieves better clustering and learns superior feature representations, thereby enhancing the overall performance of EFSS-CIL.
% \subsection{ImageNet ViT}
% \color{red} Keep class number 51, 52....,... last task 91,92,....100 \color{black}

% \subsection{MoCo-v3 ViT}

% \newpage
\subsection{Challenging Scenarios}
\subsubsection{One-shot EFSS-CIL}

\begin{figure}[t]
\begin{tikzpicture}
    % \begin{groupplot}[group style={group size=2 by 2},height=4cm,width=8cm]
    \definecolor{color0}{HTML}{DC796A} %RED
    \definecolor{color1}{HTML}{ECA576} %Orange
    \definecolor{color2}{HTML}{357FAB} %BLUE

    % \definecolor{color0}{HTML}{ecc8af} %RED
    % \definecolor{color1}{HTML}{e7ad99} %Orange
    % \definecolor{color2}{HTML}{ce796b} %BLUE

    % \definecolor{color0}{HTML}{EBA4C2} 
    % \definecolor{color1}{HTML}{ffc6d0} 
    % \definecolor{color2}{HTML}{e7ad99} 

    % \definecolor{color2}{HTML}{696969}
 .  \begin{groupplot}[group style={group size= 1 by 1, horizontal sep = 1.8cm, vertical sep = 1.5cm, },height=4cm,width=12cm]

    \nextgroupplot[xlabel={Tasks}, ylabel={Cumulative Acc.}, 
                   x label style={at={(axis description cs:0.5,0.05)},anchor=north},
                   y label style={at={(axis description cs:0.02,0.5)},anchor=north},
                   xmin=0.5, xmax=21.98, xtick={1,2,3,4,5,6,7,8,9,10,11,12,13,14,15,16,17,18,19,20,21}, 
                   xticklabels={1,2,3,4,5,6,7,8,9,10,11,12,13,14,15,16,17,18,19,20}, 
                   ymin=47.5, ymax=80, ytick={50,60,70,80}, 
                   title={\footnotesize {One Shot (ImageNet pre-trained)}},
                   title style={anchor=north, yshift=2.5ex},
                   legend pos=north west,
                   legend style={at={(0.05,1.5)},draw=black,legend columns=3, legend entries={SLCA, SLCA + Fixed Threshold, TACLE (ours)}, legend cell align={left},/tikz/every even column/.append style={column sep=0.5cm},nodes={scale=0.85, transform shape}},
                   grid=major]

        % [68.8, 63.8, 67.87, 66.7, 67.28, 63.87, 62.29, 59.5, 58.13, 57.0, 56.51, 54.3, 54.77, 55.51, 55.52, 55.48, 54.52, 55.96, 55.92, 56.02] : 59.0
        \addplot[very thick, color= color0, mark=*, mark options={solid}]  coordinates {
                (1, 68.8)
                (2, 67.87)
                (3, 66.7)
                (4, 67.28)
                (5, 63.8)
                (6, 63.87)
                (7, 62.29)
                (8, 59.5)
                (9, 58.13)
                (10, 57.0)
                (11, 56.51)
                (12, 54.3)
                (13, 54.77)
                (14, 55.51)
                (15, 55.52)
                (16, 55.48)
                (17, 54.52)
                (18, 55.96)
                (19, 55.92)
                (20, 56.02)
                    };
                   \node [right, fill=white,  inner sep=0pt] at (axis cs:20,56.02) {\color{color0} \fontsize{6}{6} \textbf{$59.48$}};
                   
        % \addlegendentry{SLCA}
        
       % [64.4, 62.2, 61.73, 61.9, 65.84, 61.27, 60.17, 56.2, 54.93, 54.68, 54.0, 53.5, 53.88, 55.8, 56.48, 57.32, 56.52, 58.2, 58.38, 59.0] : 58.32
        \addplot[very thick, color= color1, mark=*, mark options={solid}]  coordinates {
                        (1, 70.12)
                        (2, 69.21)
                        (3, 68.48)
                        (4, 67.43)
                        (5, 65.84)
                        (6, 65.27)
                        (7, 63.67)
                        (8, 61.2)
                        (9, 60.93)
                        (10, 58.68)
                        (11, 58.0)
                        (12, 56.5)
                        (13, 57.68)
                        (14, 56.8)
                        (15, 57.48)
                        (16, 57.32)
                        (17, 57.42)
                        (18, 57.1)
                        (19, 58.38)
                        (20, 57.3)
                    };
                
                    \node [right, fill=white,  inner sep=0pt] at (axis cs:20,59.0) {\color{color1} \fontsize{6}{6} \textbf{$61.23$}};
        % \addlegendentry{SLCA + Fixed Threshold}

        % [64.4, 64.8, 74.93, 71.8, 73.84, 70.07, 69.14, 67.35, 66.4, 64.04, 64.58, 65.47, 65.91, 67.14, 67.17, 67.58, 66.19, 67.38, 68.21, 68.1] : 67.725
        \addplot[very thick, color= color2, mark=*, mark options={solid}]  coordinates {
                        (1, 74.93)
                        (2, 73.84)
                        (3, 71.8)
                        (4, 70.07)
                        (5, 69.14)
                        (6, 68.21)
                        (7, 68.1)
                        (8, 67.35)
                        (9, 67.38)
                        (10, 66.4)
                        (11, 67.14)
                        (12, 67.17)
                        (13, 67.58)
                        (14, 66.19)
                        (15, 64.58)
                        (16, 65.47)
                        (17, 65.91)
                        (18, 64.04)
                        (19, 64.4)
                        (20, 64.8)
                    };
                   
                    \node [right, fill=white,  inner sep=0pt] at (axis cs:20, 64.8) {\color{color2} \fontsize{6}{6} \textbf{$67.73$}};
        % \addlegendentry{TaLE}
                
    \end{groupplot}

\end{tikzpicture}
\caption{Evaluation of One-Shot Performance on ImageNet-100 with MoCo v3 Initialization. The experiment uses 1 labeled sample and 1300 unlabeled samples per class. The 100 classes divided into 20 tasks with 5 classes per task.}
\label{fig:one_shot_imagenet100_graph}
\end{figure}
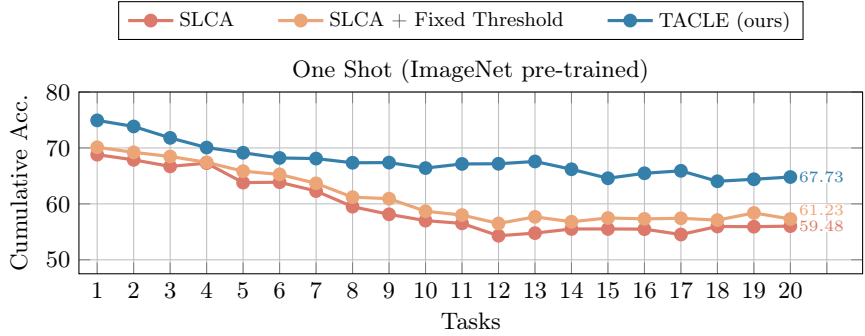

Fig. \ref{fig:one_shot_imagenet100_graph} depicts the performance of different methods in the one-shot EF-SSCIL setting for the ImageNet-Subset100 dataset. In this setting, each class has only one labeled data point along with unlabeled data, hence it is referred to as the one-shot EF-SSCIL protocol. MoCo v3 pre-trained ViT is used for weight initialization in these experiments. The ImageNet-Subset 100 dataset is divided into 20 tasks, with each task containing 5 classes. Therefore, the number of labeled and unlabeled samples per task is 5 and 6500, respectively. Our method (TACLE) achieves a $8.75\%$ higher accuracy compared to the SLCA method on this challenging setting.

\subsubsection{Imbalance EFSS-CIL}

% Define bar chart colors
\definecolor{bblue}{HTML}{4F81BD}
\definecolor{rred}{HTML}{C0504D}
\definecolor{ggreen}{HTML}{9BBB59}
\definecolor{ppurple}{HTML}{9F4C7C}
\definecolor{skin}{HTML}{F7DED0}
\definecolor{ggrey}{HTML}{C4CDDE}

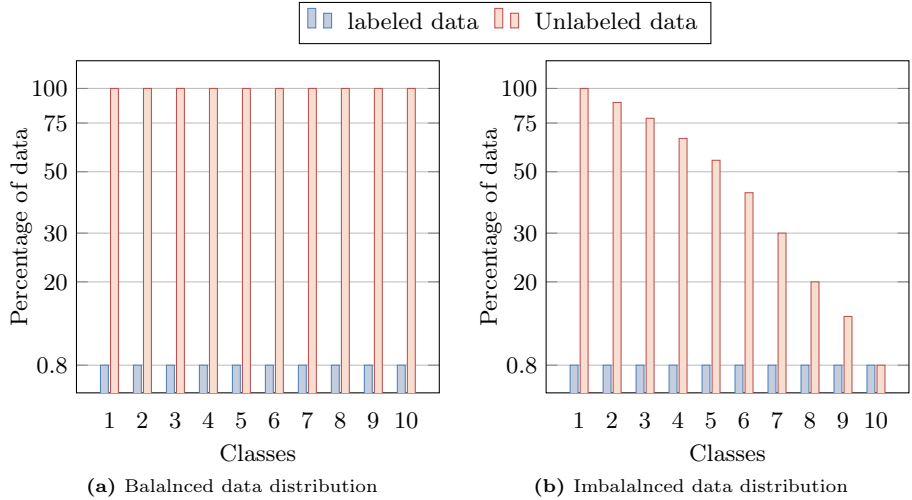
\begin{figure}[t!]
    \centering
    \begin{subfigure}{0.5\textwidth}
        \centering
        \begin{tikzpicture}
            \begin{axis}[
                width  = 1.05*\textwidth,
                height = 6cm,
                major x tick style = transparent,
                x label style={at={(axis description cs:0.5,-0.01)},anchor=north},
                y label style={at={(axis description cs:0.05,0.5)},anchor=north},
                ybar=2*\pgflinewidth,
                bar width=3pt,
                ymajorgrids = true,
                ylabel = {Percentage of data},
                xlabel = {Classes},
                xtick = data,
                scaled y ticks = false,
                scaled x ticks = false,
                ymin=2.5, 
                ymin=0,
                xmin=0,
                ymode = log,
                ytick={10,20,30,50,75,100},
                yticklabels = {0.8,20,30,50,75,100},
                legend cell align=left,
                legend style={at={(1.75,1.05)}, anchor=south east,legend columns=2, column sep=1ex}
            ]
                % \addplot[style={bblue,fill=ggrey,mark=none}]
                %     coordinates {(1, 0.8) (2,0.8) (3,0.8) (4,0.8) (5,0.8) (6,0.8) (7,0.8)(8,0.8) (9,0.8) (10,0.8)};
                    \addplot[style={bblue,fill=ggrey,mark=none}]
                    coordinates {(1, 10) (2,10) (3,10) (4,10) (5,10) (6,10) (7,10)(8,10) (9,10) (10,10)};
                \addplot[style={rred,fill=skin,mark=none}]
                    coordinates {
                    (1,100) 
                    (2,100) 
                    (3,100)
                    (4,100)
                    (5,100)
                    (6,100)
                    (7,100)
                    (8,100)
                    (9,100)
                    (10,100)};
                \legend{labeled data, Unlabeled data}
            \end{axis}
        \end{tikzpicture}
        \caption{Balalnced data distribution}
        \label{fig:balanced_dist}
    \end{subfigure}%
    \begin{subfigure}{0.5\textwidth}
        \centering
        \begin{tikzpicture}
            \begin{axis}[
                width  = 1.05*\textwidth,
                height = 6cm,
                major x tick style = transparent,
                x label style={at={(axis description cs:0.5,-0.01)},anchor=north},
                y label style={at={(axis description cs:0.05,0.5)},anchor=north},
                ybar=2*\pgflinewidth,
                bar width=3pt,
                ymajorgrids = true,
                ylabel = {Percentage of data},
                xlabel = {Classes},
                xtick = data,
                scaled y ticks = false,
                scaled x ticks = false,
                ymode = log,
                ytick={10,20,30,50,75,100},
                yticklabels = {0.8,20,30,50,75,100},
                ymin=0,
                xmin=0,
                % legend cell align=left,
                % legend style={at={(1,1.05)}, anchor=south east, column sep=1ex}
            ]
                % \addplot[style={bblue,fill=ggrey,mark=none}]
                %     coordinates {(1, 0.8) (2,0.8) (3,0.8) (4,0.8) (5,0.8) (6,0.8) (7,0.8)(8,0.8) (9,0.8) (10,0.8)};
                    \addplot[style={bblue,fill=ggrey,mark=none}]
                    coordinates {(1, 10) (2,10) (3,10) (4,10) (5,10) (6,10) (7,10)(8,10) (9,10) (10,10)};
                \addplot[style={rred,fill=skin,mark=none}]
                    coordinates {
                    (1,100) 
                    (2,89) 
                    (3,78)
                    (4,66)
                    (5,55)
                    (6,42)
                    (7,30)
                    (8,20)
                    (9,15)
                    (10,10)};
                    
                % \legend{labeled data, Unlabeled data}
            \end{axis}
        \end{tikzpicture}
        \caption{Imbalalnced data distribution}
        \label{fig:imbalanced_dist}
    \end{subfigure}
    \caption{The bar graph illustrates the data distribution for the balanced and imbalanced unlabeled data per class-wise in the CIFAR100 dataset with $0.8\%$ labeled data.}
    \label{fig:two_plots}
\end{figure}

Fig. \ref{fig:balanced_dist} illustrates the data distribution in the standard SS-CIL setting, where the unlabeled data from every class is balanced, meaning the number of samples from all classes is equal in the unlabeled data (in the standard setting, they have access to exemplars also but we are not showing for simplicity). Conversely, Fig. \ref{fig:imbalanced_dist} shows the data distribution for the imbalance EFSS-CIL proposed in the paper. In this scenario, we have a highly skewed distribution for the unlabeled data, with an imbalance ratio of 0.01, indicating that the ratio between the class with fewer samples and the class with more samples is 0.01. At every task, unlabeled data follows this imbalance (head-tail) distribution.
\subsection{Training optimization details}
During training, stage 1 for each task is trained for $10$ epochs. A learning rate schedule is employed, reducing the learning rate by a factor of $10$ after the $8^{th}$ epoch. To facilitate stable initial convergence, the network is first warmed up for a few iterations using only labeled data loss. Subsequently, unlabeled data losses are incorporated and added to the total loss function. The standard SGD optimizer with a batch size of 128 is employed for both CIFAR-10 and CIFAR-100 experiments. Due to GPU memory limitations, a reduced batch size of 64 is used for the ImageNet-subset100 experiments.
\bibliographystyle{splncs04}
\bibliography{egbib}
\end{document}